\documentclass[11pt]{article}

\PassOptionsToPackage{table}{xcolor}

\usepackage[final]{acl}

\providecommand{\linenumbers}{}

\usepackage{times}
\usepackage{latexsym}
\usepackage[T1]{fontenc}
\usepackage[utf8]{inputenc}
\usepackage{microtype}

\usepackage{amsmath}
\usepackage{amssymb}
\usepackage{booktabs}
\usepackage{multirow}
\usepackage{makecell}
\usepackage{wasysym}  
\newcommand{\cfull}{\CIRCLE}
\newcommand{\chalf}{\LEFTcircle}
\newcommand{\cnone}{\Circle}
\usepackage{graphicx}
\usepackage{xcolor}
\usepackage{algorithm}
\usepackage{algpseudocode}

\definecolor{rhoAccent}{HTML}{2F4F6F}
\definecolor{rhoComment}{HTML}{6F7480}
\definecolor{rowAlt}{HTML}{ECECEC}
\definecolor{rhoBlue}{HTML}{D6E6F2}

\newcommand{\joy}[1]{}

\usepackage{listings}
\usepackage{enumitem}

\setlist[itemize]{leftmargin=*,topsep=0pt,itemsep=1pt}
\setlist[enumerate]{leftmargin=*,topsep=0pt,itemsep=1pt}
\usepackage[most]{tcolorbox}
\tcbuselibrary{listings,breakable,skins}

\definecolor{promptBg}{HTML}{F8FAFC}     
\definecolor{promptStroke}{HTML}{2F4F6F} 

\lstdefinestyle{prompt}{%
  basicstyle=\small\ttfamily,
  columns=fullflexible,
  keepspaces=true,
  breaklines=true,
  breakatwhitespace=true,
  breakindent=1.5em,
  postbreak=\mbox{\textcolor{rhoComment}{$\hookrightarrow$}\space},
  showstringspaces=false,
  aboveskip=0pt, belowskip=0pt,
}

\newtcblisting[auto counter]{promptbox}[2][]{%
  breakable, enhanced,
  listing only,
  listing options={style=prompt},
  colback=promptBg,
  colframe=promptStroke,
  boxrule=0.6pt,
  arc=4pt, outer arc=4pt,
  left=8pt, right=7pt, top=5pt, bottom=5pt,
  boxsep=1pt,
  before skip=1.2em, after skip=0.8em,
  colbacktitle=promptStroke,
  coltitle=white,
  fonttitle=\bfseries\footnotesize,
  toptitle=2.5pt, bottomtitle=2.5pt,
  title={Listing~\thetcbcounter\quad #2},
  label={#1},
}

\algrenewcommand\algorithmiccomment[1]{\unskip\hfill\textcolor{rhoComment}{\,$\triangleright$\,\textit{\footnotesize #1}}}
\algrenewcommand\algorithmicrequire{\textcolor{rhoAccent}{\textbf{Input:}}}
\algrenewcommand\algorithmicensure{\textcolor{rhoAccent}{\textbf{Output:}}}
\algrenewcommand\algorithmicreturn{\textcolor{rhoAccent}{\textbf{return}}}
\newcommand{\rhostage}[1]{\Statex \textcolor{rhoAccent}{\rule[1pt]{6pt}{6pt}}\,\textcolor{rhoAccent}{\textsc{\textbf{#1}}}}

\title{Evolving Agents in the Dark: \\ \textsc{Retrospective Harness Optimization} via Self-Preference}

\author{
  \textbf{Wenbo Pan\textsuperscript{1}}\thanks{\;\,Corresponding author. Work done during an internship at Microsoft Research Asia.}\textbf{ \quad Shujie Liu\textsuperscript{2} \quad Chin-Yew Lin\textsuperscript{2} \quad Jingying Zeng\textsuperscript{2} \quad Xianfeng Tang\textsuperscript{2}} \\
  \textbf{Xiangyang Zhou\textsuperscript{2} \quad Yan Lu\textsuperscript{2} \quad Xiaohua Jia\textsuperscript{1}} \\[4pt]
  \textsuperscript{1}City University of Hong Kong \qquad \textsuperscript{2}Microsoft Research Asia
}

\newcommand{\method}{\textsc{RHO}}

\begin{document}
\maketitle

\begin{abstract}
AI agents rely on a \textit{harness} of skills, tools, and workflows to solve complex problems.
Continually improving this harness is essential for adapting to new tasks.
However, existing optimization methods typically require ground-truth validation sets, yet such labeled data is difficult to acquire in practical deployment settings.
To address this problem, we introduce \textit{Retrospective Harness Optimization} (\method{}), a self-supervised method that optimizes the agent harness using only past trajectories.
Specifically, \method{} selects a diverse coreset of challenging tasks from past trajectories and re-solves them in parallel.
The agent analyzes these rollouts using self-validation and self-consistency, then generates candidate harness updates and selects the most effective one by its own pairwise self-preference.
We evaluate \method{} across three diverse domains, spanning software engineering, technical work, and knowledge work.
Notably, a single optimization round improves the pass rate on SWE-Bench Pro from 59\% to 78\% without any external grading.
Furthermore, our analysis demonstrates that \method{} effectively targets prior failure modes. As a result, the optimized harness alters the agent's behavior patterns and sustains higher accuracy during long-horizon sessions.
Code is available at \url{https://github.com/wbopan/retro-harness} and the project website at \url{https://paper-rho.wenbo.io}.
\end{abstract}

\section{Introduction}

\begin{figure}[!t]
\centering
\includegraphics[width=\columnwidth]{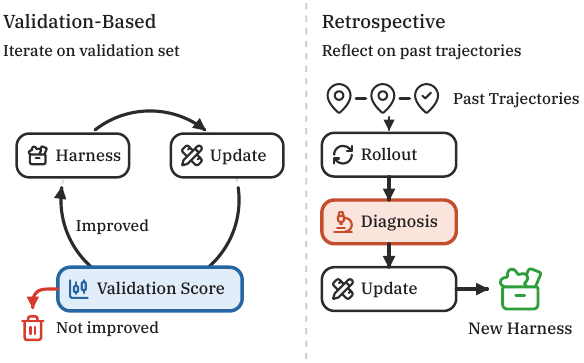}
\caption{\looseness=-1 \method{} versus validation-feedback harness optimization.
Validation-feedback methods iterate against a labeled validation set, whereas \method{} optimizes from past trajectories in a single retrospective pass with no ground-truth labels.}
\vspace{-1.5em}
\label{fig:rho-comparison}
\end{figure}

\looseness=-1 A \textit{harness} enables an AI agent to complete complex tasks by providing it with available skills, workflows, and tools.
One important research question is how to improve the harness continuously.
Specifically, after an agent is deployed, we aim to continually evolve its harness by learning from past experiences, which in turn improves its performance on future tasks.

Prior work has proposed various methods for evolving the agent harness \citep{zhou2022ape,yang2023opro,khattab2023dspy,yuksekgonul2024textgrad,agrawal2025gepa,hu2024adas,lee2026metaharness}.
However, these methods rely on scoring against a validation set to guide the improvements.
In practical deployment scenarios, it is often difficult to collect a validation set that accurately estimates the distribution of future tasks to validate the updated harness.
On the other hand, the continuous operation of an agent naturally produces a rich set of trajectories from past tasks.
This leads to our central question. \textit{Can we improve the agent harness to enhance future performance when we only have access to past trajectories?}

To address this problem, we propose \textit{Retrospective Harness Optimization} (\method{}), a self-supervised method that optimizes the harness through a retrospective analysis of past trajectories.
This method employs the agent's internal self-preference over trajectories to guide the optimization process.
Figure~\ref{fig:rho-comparison} contrasts \method{} with conventional validation-feedback optimization, which iterates against a labeled validation set.
\begin{figure*}[!t]
\centering
\includegraphics[width=\textwidth]{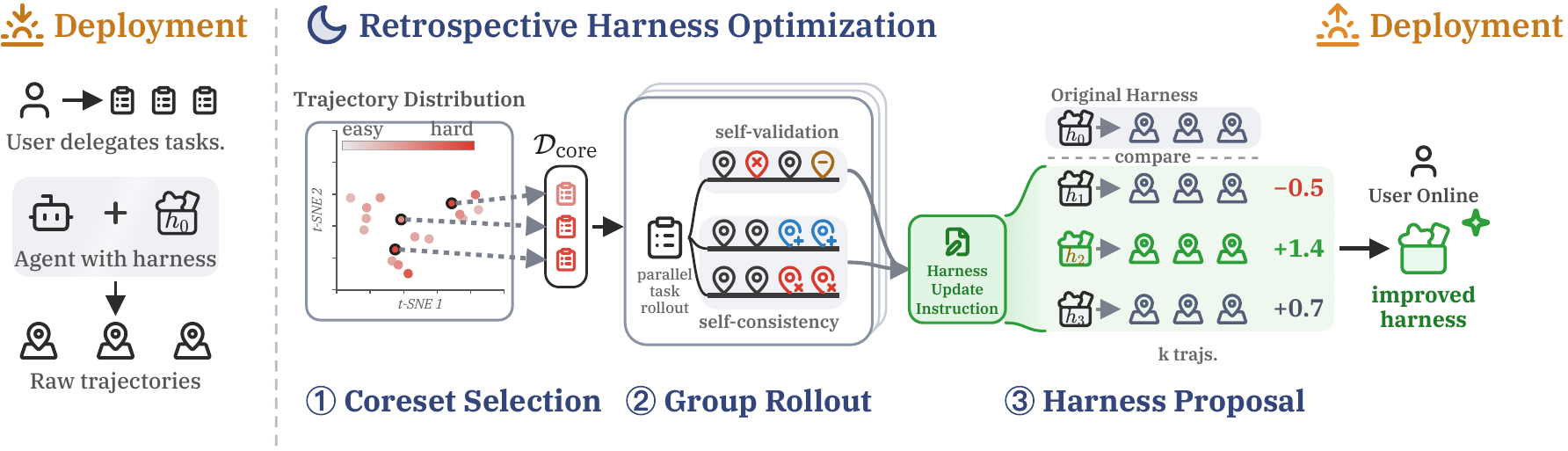}
\caption{\looseness=-1 The \method{} pipeline. \emph{Coreset Selection} picks a small, difficulty-diverse subset of past tasks with a determinantal point process (DPP). \emph{Group Rollout} re-solves each task $G$ times and diagnoses within-trajectory failures (self-validation) and cross-trajectory disagreements (self-consistency). \emph{Harness Proposal} samples $N$ candidate harnesses and keeps the one whose rollouts are most preferred over the baseline. No ground-truth labels are used.}
\vspace{-1em}
\label{fig:rho-pipeline}
\end{figure*}
\looseness=-1 Figure~\ref{fig:rho-pipeline} illustrates this process.
Specifically, given a large set of past trajectories, \method{} first selects a diverse and challenging coreset of tasks.
Then the agent re-attempts each task in the coreset multiple times to generate parallel trajectories.
Building on this, we extract two diagnostic signals, namely self-validation within a trajectory and self-consistency across parallel trajectories.
These signals are then used to instruct the generation of harness updates.
Finally, by using the agent's pairwise self-preference, we select the most promising harness from the newly generated proposals.

\looseness=-1 We evaluate the effectiveness of \method{} across three agent domains that span software engineering, technical work, and knowledge work.
\method{} consistently improves the agent's performance across all three domains.
Notably, by running a single round of retrospective harness optimization on software-engineering trajectories, we improve the pass rate on SWE-Bench Pro \citep{deng2025swebenchpro} from 59\% to 78\%, without depending on grading against a validation set.

\looseness=-1 Furthermore, we provide a detailed analysis on how the retrospective optimization process improves performance.
We observe that \method{} designs specific skills and tools targeting typical failure modes encountered in past tasks. These components reshape the agent's action patterns and help it sustain higher accuracy in long-horizon sessions.
Additionally, we quantitatively analyze the contributions of the diagnostic signals during the retrospective process.
This analysis demonstrates that each step in \method{} progressively isolates signals that contribute to performance improvements.

\looseness=-1 Our contributions are as follows:
\begin{itemize}[label=$\diamond$]
    \item We propose retrospective harness optimization, which addresses the gap of improving the full harness (including memory, context, skills, and tools) exclusively from unlabeled trajectories.
    \item We evaluate \method{} across three scenarios and show that retrospective analysis consistently outperforms straightforward experience accumulation and reaches quality comparable to validation-feedback-driven evolution with no labels at roughly a third of its compute budget.
    \item We provide a quantitative analysis of the impact of harness optimization on agent performance, showing that gathering effective improvement signals leads to targeted changes in the harness and optimizes the agent's behavior.
\end{itemize}

\section{Related Work}
\label{sec:related}

\paragraph{Harness optimization.}
Harness optimization improves an agent by editing the prompts, program parameters, or workflow code that surround a fixed model.
One line optimizes prompts or pipeline parameters against a labeled metric, spanning LLM-as-optimizer search \citep{yang2023opro}, declarative pipeline compilation \citep{khattab2023dspy}, textual-gradient updates \citep{yuksekgonul2024textgrad}, reflective prompt evolution \citep{agrawal2025gepa}, and cross-model prompt evolution \citep{wu2026mulvul}.
A more agentic line lets a meta-agent rewrite the agent's own code, where ADAS searches the space of agentic system designs \citep{hu2024adas}, Meta-Harness searches over harness code using the execution traces and scores of prior candidates \citep{lee2026metaharness}, and M$^\star$ evolves the memory system as an executable program, scoring candidate memory programs on a labeled validation set \citep{pan2026mstar}.
Although these methods differ in the surface they edit, all of them steer the search with a labeled validation metric.
\method{} departs from this paradigm, requiring no validation feedback and improving the harness in a single retrospective pass over unlabeled past trajectories.

\paragraph{Agent self-improvement.}
\looseness=-1 A second line improves agents from their own past experience, using the agent's self-judgment over trajectories in place of ground-truth labels.
Dynamic Cheatsheet maintains a self-curated memory of reusable strategies and code snippets at test time \citep{suzgun2025dynamiccheatsheet}, while ReasoningBank distills generalizable reasoning strategies from self-judged successes and failures \citep{ouyang2026reasoningbank}.
MemMA coordinates the memory cycle with multiple agents and repairs its memory bank against self-generated probe questions \citep{lin2026memma}, and Sleep-time Compute precomputes useful context offline before queries arrive \citep{lin2025sleeptime}.
Concurrent to our work, SkillOS instead trains a skill curator with reinforcement learning from outcome and judge rewards, updating a skill repository from accumulated experience \citep{ouyang2026skillos}.
These methods enrich an agent's stored memory, context, or skill list while leaving the rest of the harness untouched.
\method{} instead optimizes the full harness, including executable tools and instructions, rather than memory alone.
Appendix~\ref{app:positioning} gives a detailed comparison with related work.

\section{Problem Setting}
\label{sec:setting}

We define a \textit{harness} $h$ as a persistent collection of tools, prompts, and skills that an agent can use to solve a task.
Given a task $t$ and a harness $h$, an agent can attempt the task using a loop of reasoning, acting, and observing.
This multi-step process generates a \textit{trajectory} $\tau$, which records the information read by the agent, its chain of thought, the tools used, and the final output.
We denote this execution process with a prompted agent operation as $\tau = \mathrm{solve}(h, t)$.
As the agent executes multiple tasks, it produces a dataset of trajectories $\mathcal{D} = \{\tau_1, \tau_2, \ldots, \tau_n\}$.
These trajectories often contain instances of failure and useful insights that can be used to improve the harness.
Consequently, we ask whether the agent can retrospectively analyze past trajectories to optimize its harness and improve its future performance.
To quantify this, we define a latent \textit{utility function} $U(t, \tau)$ that measures the quality of a trajectory.
We formalize the optimization as a function $\mathrm{optimize}(h, \text{instruction})$ that returns a modified harness $h'$.
The goal is to find an optimal harness $h^\star$ that maximizes the expected utility on future tasks:
\vspace{-0.5em}
\[
    h^\star = \arg\max_{h'} \; \mathbb{E}_{t,\, \tau \sim \mathrm{solve}(h', t)}
    \left[ U(t, \tau) \right].
\]

\noindent\textbf{Problem.}
However, estimating this utility function accurately is difficult in practice.
To evaluate the true utility of a harness, we would need a representative validation set of future tasks and a mechanism to calculate the success rate of the agent using this specific harness.
In our setting, the function $U$ is latent and cannot be directly observed.

\noindent\textbf{Our Approach.}
Because the utility $U$ is latent, we cannot directly optimize it.
Instead, we substitute this latent utility with a self-preference estimator.
Specifically, we instruct the agent to compare multiple trajectories on the same task to compute a self-preference ranking.
We define a ranking function as $(\text{rank}, \text{rationale}) = \mathrm{rank}(t, \tau_1, \tau_2, \ldots, \tau_m)$.
This function yields a preference ordering over the given trajectories and provides a rationale that explains why the agent prefers certain executions over others.
The next section details how we organize the operations of solving, ranking, and optimizing to improve the latent harness utility.
\begin{algorithm}[t]
\caption{\looseness=-1 A single round of \method{}. One backbone instantiates every operator (the difficulty $\mathrm{judge}$, $\mathrm{solve}$, $\mathrm{optimize}$, and $\mathrm{rank}$), differing in its inputs and consulting no ground-truth label.}
\label{alg:rho}
\footnotesize
\begin{algorithmic}[1]
\Require past trajectories $\mathcal{D}{=}\{(t_i,\tau_i)\}_i$ and harness $h_0$, with coreset size $k$, group size $G$, candidate count $N$, and DPP weight $\theta$
\Ensure  updated harness $h^\star$
\rhostage{Stage 1 \, Coreset Selection}
\State $r_i \gets \mathrm{judge}(t_i,\tau_i)\ \ \forall\,(t_i,\tau_i)\in\mathcal{D}$
\State $\mathcal{D}_{\mathrm{core}} \gets \textsc{DPP-Greedy}\bigl(\{(t_i,r_i)\};\,\theta,k\bigr)$
\rhostage{Stage 2 \, Group Rollout}
\For{$t \in \mathcal{D}_{\mathrm{core}}$ \textbf{in parallel}}
  \State $\{\tau_{t,g}\}_{g=1}^{G} \gets \mathrm{solve}(h_0,t)$ \Comment{$k{\times}G$ rollouts}
  \State $\tau_t^{(0)} \gets \tau_{t,1}$ \Comment{\emph{fixed} baseline rollout}
  \State $I_t \gets \mathrm{rank}_{\mathrm{val}}\bigl(t,\{\tau_{t,g}\}\bigr)\cup\mathrm{rank}_{\mathrm{con}}\bigl(t,\{\tau_{t,g}\}\bigr)$
\EndFor
\State $I \gets \bigcup_{t\in\mathcal{D}_{\mathrm{core}}} I_t$
\rhostage{Stage 3 \, Best-of-$N$ Harness Proposal}
\For{$j = 1,\dots,N$ \textbf{in parallel}}
  \State $h_j \gets \mathrm{optimize}\bigl(h_0,\,I\bigr)$
  \State $\tau_t^{(j)} \gets \mathrm{solve}(h_j,t)\ \ \forall\,t\in\mathcal{D}_{\mathrm{core}}$ \Comment{$N{\times}k$ re-solves}
  \State $S_j \gets \tfrac{1}{k}\sum_{t\in\mathcal{D}_{\mathrm{core}}} \mathrm{rank}\bigl(t,\,\tau_t^{(j)},\,\tau_t^{(0)}\bigr)$
\EndFor
\State $j^\star \gets \arg\max_{j} S_j$
\State \Return $h_{j^\star}$ \textbf{if} $S_{j^\star}>0$, \textbf{else} $h_0$
\end{algorithmic}
\end{algorithm}

\section{Retrospective Harness Optimization}

\looseness=-1 We propose \method{}, a self-supervised method that improves a harness using only past trajectories.
Specifically, our pipeline (Figure~\ref{fig:rho-pipeline}) consists of three stages, namely coreset selection, group rollout, and best-of-$N$ harness proposal.
First, we select a representative subset of past tasks to define the optimization target.
Next, we sample a group of parallel rollouts for each task in this coreset and extract harness improvement signals from them.
Finally, we generate $N$ candidate harnesses based on these signals and retain the most preferred one using pairwise self-preference.
The full algorithm is detailed in Algorithm~\ref{alg:rho}.

\subsection{Coreset Selection}

Given a large set of past trajectories, we need to extract the most critical signals to guide the harness optimization.
Optimizing the harness on every individual trajectory is computationally prohibitive, and it further risks diluting important signals with trivial ones.
To address this, we first select a coreset $\mathcal{D}_{\mathrm{core}}$ from the full set $\mathcal{D}$ to represent the trajectories that require optimization the most, echoing trajectory-based data valuation that scores training samples by their influence on optimization \citep{xu2025liveval}.
Specifically, we require the coreset to capture both challenging and diverse scenarios.
This requirement encourages our optimization to cover a wide range of failure modes when addressing the most difficult problems.
To accomplish this, we introduce a Determinantal Point Process (DPP) kernel \citep{kulesza2012dpp} to rank all past trajectories by difficulty while satisfying a diversity constraint.
In practice, we employ a language model judge to analyze every trajectory $\tau_i$ and extract a difficulty score $r_i$ alongside a textual description.
This description details the specific challenges of the problem and potential failure modes.
We then compute the embedding of this description and use the cosine similarity between embeddings as the similarity metric $S_{i,j}$ for any two trajectories $\tau_i$ and $\tau_j$.
By considering both the difficulty scores $r$ and the trajectory similarity matrix $S$, we construct an L-ensemble kernel matrix
\[
    L = \mathrm{diag}( \widetilde{r})\,S\,\mathrm{diag}( \widetilde{r}),
\]
where  $\widetilde{r}_i$ is a scaled version of the trajectory's difficulty score $r_i$:
\[
\widetilde{r}_i = \big(\max(r_i,\epsilon)\,/\,\max_j \max(r_j,\epsilon)\big)^{\alpha},
\]
\[
\alpha = \theta/\big(2(1-\theta)\big).
\]
With this kernel function $L$, DPP selects a subset $Y$ with probability proportional to the kernel determinant $\det(L_Y)$, using parameter $\theta$ to adjust the relative importance of difficulty and diversity via $\alpha$. With $\theta = 1$, the trajectories are ranked purely by difficulty and $\theta = 0$ (uniform weights) ranked purely by similarity diversity.
Using $\theta = 0.7$, we select $k$ trajectories into a coreset $\mathcal{D}_{\mathrm{core}}$ that covers difficult, diverse failure modes for later stages.

\subsection{Group Rollout}

Inspired by previous work that computes advantages relative to a group of sampled rollouts, using the group as a baseline in place of a learned value model \citep{shao2024deepseekmath}, we generate a set of trajectories by running $G$ parallel agent solves on each coreset task.
Subsequently, the agent compares these group trajectories to identify underperforming runs.
The agent then uses contrastive signals within the group to formulate instructions for optimizing the harness.
Specifically, we perform this self-preference analysis along two dimensions.
\begin{itemize}
    \item \looseness=-1 \textbf{Self-validation} ($\mathrm{rank}_{\mathrm{val}}$)\textbf{.} This dimension examines the correctness of the agent within each trajectory.
    The agent inspects each trajectory against the required task and environment observations to determine whether the objective is efficiently achieved, exploiting the partial ability of models to recognize the limits of their own knowledge \citep{pan2025refusal}.
    During this process, it flags incorrect tool invocations, false assumptions, and premature stopping. These flagged aspects are then extracted as areas of improvement for the relatively underperforming runs.
    \item \looseness=-1 \textbf{Self-consistency} ($\mathrm{rank}_{\mathrm{con}}$)\textbf{.} This dimension examines whether the behavior of the agent remains consistent across different trajectories.
    Because low self-consistency typically indicates high uncertainty \citep{wang2022selfconsistency,farquhar2024semantic}, we instruct the agent to analyze contradictions among trajectories.
    The agent identifies consequential disagreements, such as divergent plans, tool sequences, or final answers, and generates optimization instructions to encourage more consistent behavior.
\end{itemize}
These $\mathrm{rank}_{\mathrm{val}}$ and $\mathrm{rank}_{\mathrm{con}}$ analyses yield structured evaluations in JSON format, and for each task their union forms the improvement instruction $I_t = \mathrm{rank}_{\mathrm{val}}\bigl(t, \{\tau_{t,g}\}\bigr) \cup \mathrm{rank}_{\mathrm{con}}\bigl(t, \{\tau_{t,g}\}\bigr)$.
As a result, we merge $\{I_t\}$ across all tasks in the coreset to form the final harness improvement instructions.

\begin{table*}[t]
\caption{Held-out pass rate after harness optimization. The Architecture column indicates which harness surface each method edits. $\Delta$ is the absolute change over Vanilla Codex on the same held-out split.}
\label{tab:main}
\centering
\resizebox{\textwidth}{!}{%
\begin{tabular}{llcccccc}
\toprule
& \textbf{Harness} & \multicolumn{2}{c}{\textbf{SWE-Bench Pro}} & \multicolumn{2}{c}{\textbf{Terminal-Bench~2}} & \multicolumn{2}{c}{\textbf{GAIA-2}} \\
\cmidrule(lr){3-4} \cmidrule(lr){5-6} \cmidrule(lr){7-8}
\textbf{Method} & \textbf{Architecture} & \textbf{Pass} & $\boldsymbol{\Delta}$ & \textbf{Pass} & $\boldsymbol{\Delta}$ & \textbf{Pass} & $\boldsymbol{\Delta}$ \\
\midrule
\rowcolor{rowAlt}
Vanilla Codex & None & 0.59 & n/a & 0.71 & n/a & 0.29 & n/a \\
\midrule
\rowcolor{white}
Dynamic Cheatsheet \citep{suzgun2025dynamiccheatsheet} & Skills & 0.62 & $+0.03$ & 0.73 & $+0.02$ & 0.30 & $+0.01$ \\
\rowcolor{rowAlt}
ReasoningBank \citep{ouyang2026reasoningbank} & Memory & 0.61 & $+0.02$ & 0.73 & $+0.02$ & 0.28 & $-0.01$ \\
\rowcolor{white}
Sleep-time Compute \citep{lin2025sleeptime} & Memory & 0.64 & $+0.05$ & 0.73 & $+0.02$ & 0.32 & $+0.03$ \\
\midrule
\cellcolor{rhoBlue}\method{} & \cellcolor{rhoBlue}Skills+Tools & \cellcolor{rhoBlue}\textbf{0.78} & \cellcolor{rhoBlue}$\mathbf{+0.19}$ & \cellcolor{rhoBlue}\textbf{0.76} & \cellcolor{rhoBlue}$\mathbf{+0.05}$ & \cellcolor{rhoBlue}\textbf{0.37} & \cellcolor{rhoBlue}$\mathbf{+0.08}$ \\
\bottomrule
\end{tabular}%
}
\vspace{-1em}
\end{table*}

\subsection{Best-of-$N$ Harness Proposal}

After obtaining the improvement instructions, we optimize the harness by providing these instructions to the agent.
However, as observed in prior studies on agent evolution \citep{agrawal2025gepa,hu2024adas,lee2026metaharness}, harness optimization is inherently stochastic and may not reliably improve performance even with valid input signals.
To mitigate this limitation, we sample harness proposals in parallel and filter them using agent self-preference.
This selection is designed to favor candidates whose improvements generalize to future tasks.
Specifically, we execute $N$ parallel optimization calls to generate $N$ candidate harnesses, denoted as $h_1$ to $h_N$.
Following this step, we use these candidates to obtain $N$ sets of new trajectories on the $k$ coreset tasks.
For every coreset task, we then compute an agent preference score by ranking the new trajectory from each candidate harness against the old trajectory from the original harness.
We aggregate these scores across the coreset to determine the relative advantage score of each candidate:
\[
    S_j = \frac{1}{|\mathcal{D}_{\mathrm{core}}|}
    \sum_{t \in \mathcal{D}_{\mathrm{core}}}
    \mathrm{rank}\!\left(t, \tau_{t}^{(j)}, \tau_{t}^{(0)}\right),
\]
where $\tau_{t}^{(0)}$ is the original harness trajectory for task $t$.
Finally, we return the candidate harness with the maximum relative advantage to replace the original one.
We accept this update only if its score is strictly greater than zero ($S_j > 0$).

\section{Experiments and Results}
\begin{figure*}[!t]
\centering
\includegraphics[width=\textwidth]{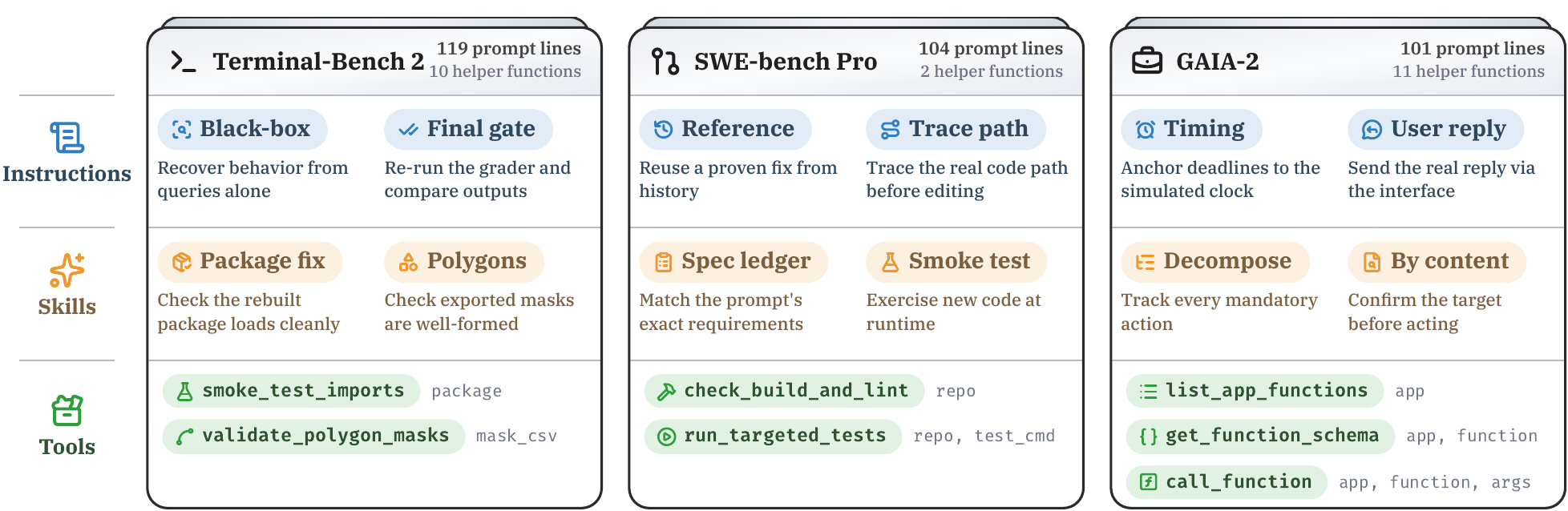}
\caption{\looseness=-1 Highest-scoring harness produced by \method{} on each benchmark. \emph{Instructions} are task-agnostic procedural rules, \emph{Skills} record grader or environment idiosyncrasies that previously caused failures, and \emph{Tools} are executable scripts. Items shown are representative, and the full verbatim contents of each harness are in Appendix~\ref{app:artifacts}.}
\label{fig:harness-artifacts}
\vspace{-1em}
\end{figure*}

\noindent\textbf{Setup.}
We use the Codex agent \citep{openai2025codex} as the base harness for retrospective optimization.
Specifically, this agent uses GPT-5.5 \citep{openai2026gpt55} configured with high reasoning effort.
When we invoke Codex to solve a task, we construct the harness as a configurable workspace folder.
This folder contains executable scripts as tools, along with text files for skills and instructions.
For all our experiments, we set the coreset size $k$ to 10.
In addition, we use 3 for both parallel trajectory sampling and harness proposals.
To measure the improvement, we report the pass rate on the held-out test set using both the vanilla Codex harness and the optimized one.

\noindent\textbf{Data.}
We collect past trajectories from existing benchmark datasets.
Specifically, we divide the original benchmarks into a trajectory set and a test set.
We then run the vanilla Codex agent on the trajectory set to generate the required trajectories for \method{}.
Building on this, we evaluate \method{} on SWE-Bench Pro, Terminal-Bench~2, and \mbox{GAIA-2}.
SWE-Bench Pro contains long-horizon software-engineering tasks requiring repository-level reasoning and multi-file edits \citep{deng2025swebenchpro}.
Terminal-Bench~2 contains command-line tasks with executable graders \citep{terminalbench2}.
\mbox{GAIA-2} evaluates LLM agents in dynamic, asynchronous environments for knowledge work \citep{froger2026gaia2}.
As a result, these three benchmarks cover a wide range of task types across software engineering, technical work, and knowledge work.
During optimization \method{} consumes only past trajectories and its own judgments; graders are used solely for held-out evaluation.
We provide detailed information about the benchmarks and data splits in Appendix~\ref{app:datasets}.

\subsection{Comparison with Feedback-Free Baselines}

We compare \method{} against three competitive harness optimization methods that do not require validation feedback.
For the baselines, \textit{Dynamic Cheatsheet} maintains a running record of useful facts and procedures \citep{suzgun2025dynamiccheatsheet}.
\textit{ReasoningBank} stores reusable reasoning patterns and retrieves the top-$k$ relevant entries at inference time \citep{ouyang2026reasoningbank}.
Similarly, \textit{Sleep-time Compute} preprocesses past traces offline into compact notes, which are then prepended to the agent's context \citep{lin2025sleeptime}.
We adapt each baseline to our datasets and agent setting while holding the total agent-call budget approximately fixed to ensure a fair comparison.
Detailed adaptation procedures are provided in Appendix~\ref{app:baselines}.

As Table~\ref{tab:main} shows, \method{} delivers consistent improvements across all three benchmarks, whereas the baselines do not.
Most notably, we achieve an absolute improvement of 19\% on SWE-Bench Pro without relying on any validation-based grading.
We attribute this advantage to the more flexible harness optimization that \method{} enables.
Specifically, the agent can create new tools, skills, and instructions for the harness, whereas previous methods focus predominantly on memory systems or text-based skills.
Furthermore, the use of self-preference may contribute to the consistency of these harness improvements.
In contrast, the performance gains of the baseline methods tend to be smaller and vary across different datasets.
In the next section (Section~\ref{sec:case-study}), we examine how \method{} modifies the harness to improve the agent.

\subsection{What the Optimized Harness Contains}
\label{sec:case-study}

Figure~\ref{fig:harness-artifacts} summarizes and interprets the new harness contents generated after \method{} optimization.
In our work, the harness is materialized as a directory containing markdown files for instructions and skills, as well as executable scripts for tools.

Across all three benchmarks, \method{} adds multiple new skills and tools to the harness.
Many of these new additions address typical failure modes encountered by the original harness.
For example, in SWE-Bench Pro, the agent learns that the Go toolchain resides at a non-standard location outside the default path.
It also discovers that Python cache directories must be stripped before producing the final diff, as failing to do so often prevents patches from applying cleanly.
To address these, the agent adds a new \texttt{check\_build\_and\_lint} tool that locates non-standard toolchains and flags the generated artifacts that must be kept out of the patch, fixing the diff-hygiene procedures the original trajectories repeatedly missed.
These examples illustrate how \method{} identifies useful tools and skills across diverse scenarios by analyzing past failures.

\subsection{Comparison with Validation-Feedback Optimization}

We next compare \method{} against Meta-Harness \citep{lee2026metaharness}.
Meta-Harness is a validation-feedback optimizer: a proposer emits harness edits, each candidate is graded on a labeled search set with the ground-truth grader, and the search keeps the harnesses that score best on that set.
To maintain a fair comparison, we use the same Codex agent as the Meta-Harness proposer and solver.
Unlike \method{}, Meta-Harness requires held-out labels, and because it evolves over multiple rounds, it consumes more agent calls.
Therefore, we evaluate it at a single round to match our compute budget, as well as in an extended-budget setting of ten rounds.

\begin{table}[t]
\caption{\method{} versus Meta-Harness, a validation-feedback optimizer, on SWE-Bench Pro. \emph{Agent calls} is the optimization-time budget relative to one \method{} run. Held-out pass rate on the same split as Table~\ref{tab:main}.}
\label{tab:meta-harness}
\centering
\footnotesize
\setlength{\tabcolsep}{4pt}
\resizebox{\columnwidth}{!}{%
\begin{tabular}{lccc}
\toprule
\textbf{Method} & \shortstack{\textbf{Val.}\\\textbf{labels}} & \shortstack{\textbf{Agent}\\\textbf{calls}} & \shortstack{\textbf{SWE-Bench}\\\textbf{Pro}} \\
\midrule
\rowcolor{rhoBlue}
\method{} & none & 103 ($1.0\times$) & 0.78 \\
\rowcolor{white}
Meta-Harness (1 round) & required & 41 ($0.4\times$) & 0.62 \\
\rowcolor{rowAlt}
Meta-Harness (10 rounds) & required & 320 ($3.1\times$) & \textbf{0.80} \\
\bottomrule
\end{tabular}%
}
\vspace{-1.5em}
\end{table}

At the matched single-round budget, Meta-Harness selects its best candidate using validation scores but achieves only a 0.62 pass rate on SWE-Bench Pro.
This is substantially lower than the 0.78 pass rate achieved by \method{}.
Scaling Meta-Harness to a 10-round setting increases its performance ceiling to 0.80 on SWE-Bench Pro.
However, this higher performance requires roughly three times the optimization-phase compute of \method{} (320 vs.\ 103 agent calls) and, more importantly, still depends on held-out labels that \method{} does not use.

\begin{figure*}[!t]
\centering
\includegraphics[width=\textwidth]{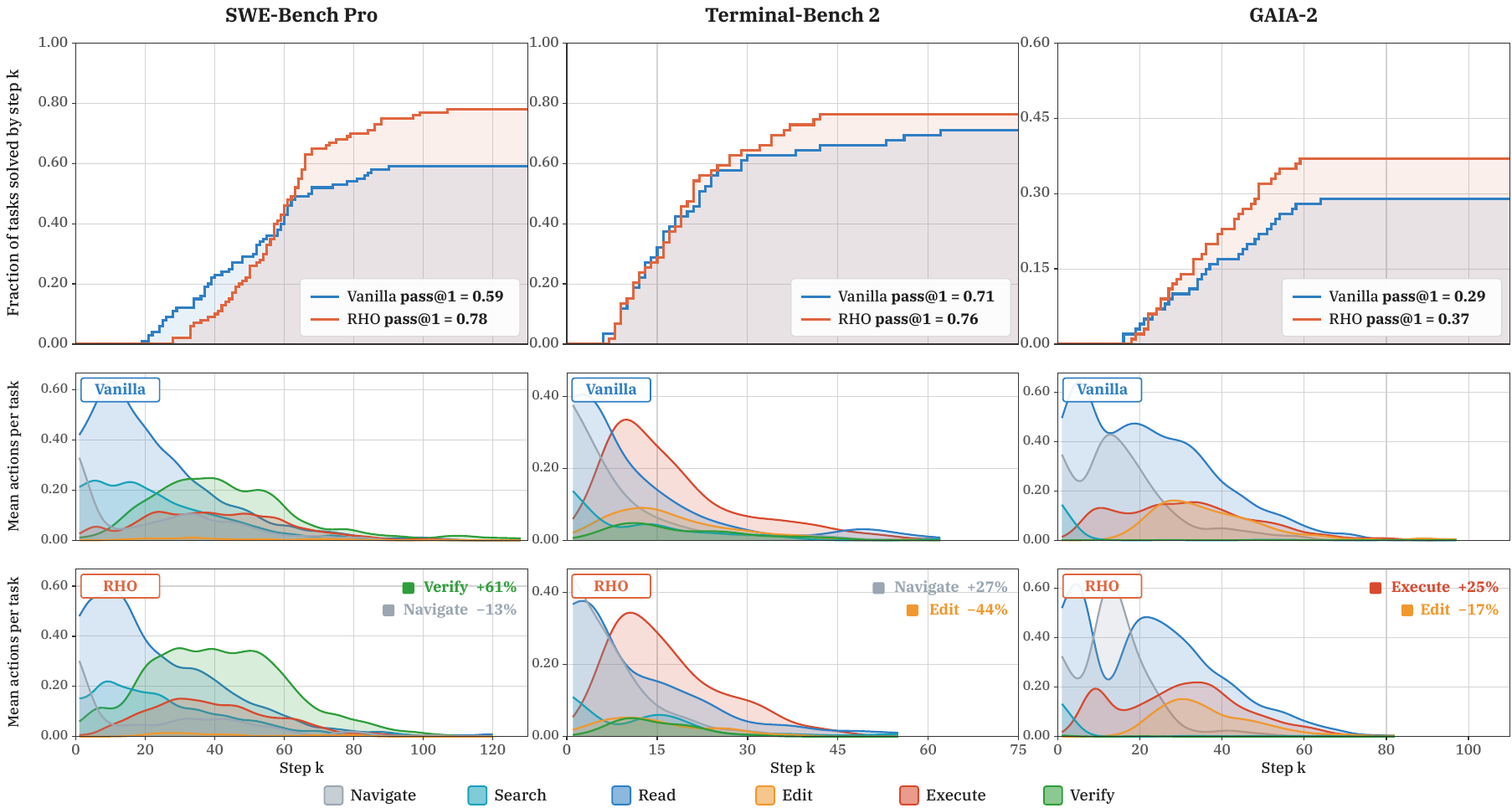}
\caption{Behavior shift after \method{}. \method{} sustains longer working sessions and shifts the agent's per-step action mix toward verification on SWE-Bench Pro, and toward execution on Terminal-Bench~2 and \mbox{GAIA-2}.}
\vspace{-1em}
\label{fig:steps}
\end{figure*}

\section{Discussion}


\subsection{How does agent behavior change after optimization?}
\label{sec:disc-behavior}

\looseness=-1 Although \method{} creates new skills and tools for the agent, it is not immediately apparent through which mechanisms these updates enable the agent to perform better on future tasks.
To examine this, we visualize the frequency of tool calls and the cumulative success rate with respect to the number of steps taken by the agent.
Specifically, Figure~\ref{fig:steps} plots the cumulative fraction of held-out tasks resolved within a given number of agent steps, and it shows how the action mix of the agent changes over time.
We observe that the performance improvements across all three datasets primarily originate from higher success rates on tasks requiring long horizons.
In contrast, the gains concentrate on long-horizon tasks rather than those completed in fewer steps, most clearly on SWE-Bench Pro.
Furthermore, the optimization process changes the working patterns of the agent.
Consequently, the optimized agent shifts toward relying more heavily on specific types of actions.
For example, on SWE-Bench Pro, the agent verifies its work much more frequently.
This proactive verification appears to account for a large portion of the performance gains on long-horizon tasks.
On Terminal-Bench~2 and \mbox{GAIA-2}, the agent increases its accuracy by actively applying newly developed tools.
The newly added tools are invoked on nearly every held-out task, and on GAIA-2 the new environment helper runs about twenty times per task.
Terminal-Bench~2 is an instructive exception---there the generated scripts are never executed, and the gain instead comes from the procedural playbook written into the instructions, showing that \method{} adapts the harness surface to each environment rather than always adding executable tools.
Complementary to our action-level statistics, token-level attribution over long reasoning chains offers a finer-grained lens on long-horizon agent behavior \citep{pan2026flashtrace}.

\subsection{How does coreset selection shape capability evolution?}
\label{sec:disc-coreset}

\begin{figure}[t]
\centering
\includegraphics[width=\columnwidth]{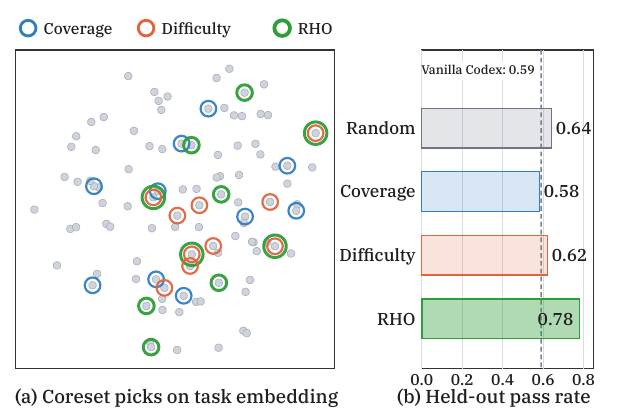}
\caption{Coreset selection on SWE-Bench Pro. \emph{(a)} Where each selector's picks land on the task embedding, with coverage spreading out, difficulty clustering, and \method{}'s DPP balancing both. \emph{(b)} Held-out pass rate of the harness optimized from each coreset. Difficulty or diversity alone trails even random sampling, and only the DPP's combination reaches the top gain.}
\label{fig:coreset-schematic}
\vspace{-1.5em}
\end{figure}

\begin{table}[t]
\caption{Best-of-$N$ harness proposal vs.\ a single sampled candidate. Held-out pass rate of the $N{=}3$ candidates. \emph{Mean} is the expected score under uniform random selection, and \emph{Chosen} is the candidate \method{} deploys.}
\label{tab:bestofn}
\centering
\footnotesize
\setlength{\tabcolsep}{4pt}
\sbox0{%
\begin{tabular}{lcccc}
\toprule
\textbf{Dataset} & \textbf{Mean} & \textbf{Chosen} & \textbf{Std} & \textbf{Lowest} \\
\midrule
\rowcolor{rowAlt}
SWE-Bench Pro & 0.79 & 0.78 & 0.06 & 0.73 \\
\rowcolor{white}
Terminal-Bench~2 & 0.74 & \textbf{0.76} & 0.03 & 0.71 \\
\rowcolor{rowAlt}
GAIA-2 & 0.34 & \textbf{0.37} & 0.03 & 0.32 \\
\bottomrule
\end{tabular}%
}
\ifdim\wd0>\columnwidth
  \resizebox{\columnwidth}{!}{\usebox0}%
\else
  \usebox0%
\fi
\end{table}

\begin{table}[t]
\caption{Ablation of the diagnosis step across all three benchmarks. Each row re-renders full \method{}'s diagnoses with one cue blanked, holding the proposal agent's prompt fixed (\S\ref{sec:disc-diagnosis}). \emph{Raw trajectory} skips diagnosis and shows the proposal agent the raw trajectories directly. Cells report held-out pass rate.}
\label{tab:diagnosis}
\centering
\footnotesize
\setlength{\tabcolsep}{4pt}
\sbox0{%
\begin{tabular}{lccc}
\toprule
\textbf{Variant} & \textbf{SWE Pro} & \textbf{TB~2} & \textbf{GAIA-2} \\
\midrule
\rowcolor{rowAlt}
Full diagnosis & \textbf{0.78} & \textbf{0.76} & \textbf{0.37} \\
\rowcolor{white}
\;\; $-$ self-consistency & 0.56 & 0.75 & 0.27 \\
\rowcolor{rowAlt}
\;\; $-$ self-validation & 0.70 & 0.73 & 0.30 \\
\rowcolor{white}
Raw trajectory & 0.60 & 0.75 & 0.29 \\
\bottomrule
\end{tabular}%
}
\ifdim\wd0>\columnwidth
  \resizebox{\columnwidth}{!}{\usebox0}%
\else
  \usebox0%
\fi
\vspace{-1.5em}
\end{table}

\looseness=-1 We investigate how coreset selection influences the optimization process.
To this end, we compare our DPP-based selection against several ablation strategies.
These variants include selecting tasks solely by difficulty ($\theta=1$), selecting tasks purely to maximize coverage ($\theta=0$), and sampling trajectories randomly.
In addition, we measure the final performance of the optimized harness under each selection strategy.
As the t-SNE projection of task embeddings in Figure~\ref{fig:coreset-schematic} reveals, selecting tasks exclusively by difficulty causes the chosen samples to cluster in a narrow region of the task distribution.
This clustering occurs because the language model judges certain types of tasks as inherently more difficult, and consequently it fails to include other task types in the coreset.
As a result, this strategy yields no meaningful performance improvement after optimization.
Similarly, optimizing solely for coverage also produces suboptimal results.
In contrast, random sampling can occasionally select a trajectory that proves useful for optimization.
These findings suggest that a coreset selection strategy balancing both difficulty and diversity is vital for providing the proper signals to guide harness optimization.

\subsection{Does \method{} produce consistent harness updates?}
\label{sec:disc-consistency}

Because \method{} operates without ground-truth labels for reference, we evaluate whether its optimization outputs are consistent across runs.
This consistency is closely tied to the overall effectiveness of our best-of-$N$ harness proposal.
We examine whether the selection strategy reliably identifies the candidate harness that performs best on downstream tasks.
In this experiment, we measure the test scores of all three generated candidate harnesses rather than just the most preferred one.
Here Table~\ref{tab:bestofn} shows that the generated harnesses exhibit only moderate variance.
Notably, even the lowest-scoring candidate still meaningfully improves agent performance over the baseline.
Furthermore, the best-of-$N$ selection prevents the deployment of poorly performing harnesses.
Specifically, the chosen harness scores higher than the worst candidate across all three benchmarks.
At the same time, the most preferred harness does not invariably coincide with the highest-scoring candidate on the test set, though the selection consistently avoids the worst candidate.
Comparing each candidate's preference score $S_j$ with its held-out pass rate further shows that self-preference is a conservative selector rather than a perfect ranker: on SWE-Bench Pro a less-preferred candidate in fact reached 0.85, above the chosen one's 0.78.
What the mechanism reliably guarantees is safety rather than perfect ranking.
Because an update is deployed only when the agent prefers it over the current harness ($S_j > 0$), the deployed harness never fell below the vanilla baseline on any benchmark; every candidate beats vanilla on SWE-Bench Pro and GAIA-2, and on Terminal-Bench~2 the worst candidate matches vanilla (0.71) while the chosen one reaches 0.76.

\subsection{How much does retrospective analysis contribute?}
\label{sec:disc-diagnosis}

We analyze the contribution of the two signals extracted during retrospective analysis, namely self-validation and self-consistency.
We compare these explicit signals against a more direct approach that provides raw trajectories during optimization and skips the explicit retrospective analysis step.
In addition, we study the individual contribution of each diagnostic signal to the final performance.

To investigate the necessity of the group rollout and diagnostic stages, we conduct an ablation study.
Specifically, we remove the self-validation and self-consistency signals independently, and we subsequently rerun the optimization and evaluation procedures.
We also introduce a raw trajectory baseline.
This baseline bypasses the separate ranking analysis. Instead, it provides the original trajectories directly to the optimization step, asking the agent to analyze the trajectory and propose improvements in a single pass.
Table~\ref{tab:diagnosis} shows that removing either the self-consistency or the self-validation signal consistently degrades final performance across benchmarks.
This result indicates that both signals are highly important for optimizing the harness.
Additionally, full diagnosis outperforms the simplified raw-trajectory baseline on all three benchmarks.
The raw-trajectory baseline grants the proposal agent the same editable surface as \method{}---tools, skills, and instructions---so the gap to full diagnosis (0.60 vs.\ 0.78 on SWE-Bench Pro) reflects the value of the structured retrospective analysis rather than a broader editing surface.
This result suggests that although single-pass trajectory analysis is feasible, the explicit self-validation and self-consistency signals are essential rather than incidental, yielding more reliable improvements across datasets.
Finally, Table~\ref{tab:diagnosis} is a controlled ablation that removes one diagnosis cue while holding the rest of the pipeline on a fixed path ($n{=}1$), so it measures the marginal contribution of each cue; it should not be read as the end-to-end effect of disabling a signal.
To measure run-to-run robustness, we additionally reran the full pipeline end to end twice per ablated strategy, with independently generated candidate sets: every rerun configuration stayed within $\pm$0.01--0.03 of its mean ($0.79 \pm 0.01$ without self-consistency and $0.815 \pm 0.025$ without self-validation on SWE-Bench Pro, $n{=}2$), and in no rerun did performance drop to the controlled-ablation level.
With $n{=}2$ this is indicative rather than conclusive, but it shows that the best-of-$N$ proposal and the remaining signal compensate when a cue is missing, so removing one component does not destabilize the pipeline.

\section{Conclusion}

We introduced \method{}, which reframes harness improvement as a retrospective process that an agent can run on its own past experience, rather than a search guided by external ground-truth feedback.
The central idea is that an agent's own trajectories already contain the signal needed to improve it, since re-solving past tasks and comparing the outcomes exposes where the harness fails and what would fix it.
Across all three domains (software engineering, technical work, and knowledge work), this self-supervised loop yields consistent held-out gains and reshapes how the agent works.
These results are established on a Codex-style CLI agent over a single retrospective round; extending \method{} to other backbones and agent frameworks, and to iterative multi-pass optimization, is natural future work.
We see \method{} as a step toward agents that keep improving from the experience they accumulate in deployment, where labeled data is rare.

\section*{Limitations}

In this paper we introduce \method{}, a self-supervised method that improves an agent's harness from its own past trajectories without any external grading.
However, operating without ground-truth feedback carries several limitations.
First, group rollout replays each coreset task several times, which assumes environments that reset cleanly and tolerate repeated attempts, leaving one-shot or irreversible tasks outside the setting \method{} targets.
Second, \method{} presumes that a meaningful portion of the agent's competence is mediated by an editable harness of prompts, skills, and tools; our experiments span software engineering, technical work, and knowledge work, and extending \method{} to domains with different harness surfaces, task fingerprints, and rollout budgets remains future work.
Third, all experiments use a single backbone and agent framework (a Codex-style CLI agent running GPT-5.5). The method's signals come only from the agent's own trajectories and pairwise judgments, so it does not inherently depend on this particular backbone, but verifying that it transfers to other backbones and agent frameworks requires further experiments that we leave to future work.
Fourth, we evaluate a single retrospective optimization round and make no claim that gains compound or even remain monotonic across multiple rounds; studying iterative, multi-pass \method{} over sequential trajectory batches is future work.
Fifth, the optimized harnesses are adapted to their own benchmark environment (for example, a repair-verification tool for SWE-Bench Pro versus a procedural playbook for Terminal-Bench~2) rather than forming one universal harness; we leave a formal cross-benchmark transfer study to future work.
Sixth, \method{} trusts past trajectories as its only optimization input; in open environments trajectories can embed adversarial content injected mid-task \citep{liu2026webtrap}, and harness updates distilled from compromised trajectories could entrench such behavior, reinforcing the audit practices in the Ethics Statement.

\section*{Ethics Statement}

\method{} modifies persistent agent behavior from model-generated judgments.
This can amplify mistaken preferences, unsafe procedures, or biased behavioral rules if the evaluator prefers them.
Deployments should keep full audit logs, require human approval for sensitive harness edits, and use domain-specific safety checks before applying accepted harnesses to high-impact tasks.

\section*{Reproducibility Statement}

Every run persists prompts, completions, trajectories, diagnoses, candidate harnesses, harness diffs, configs, scores, run metadata, and held-out reports.
The numbers in this draft are direct reads from recorded run reports.
A detailed comparison with related work is given in Appendix~\ref{app:positioning}, prompts are listed in Appendix~\ref{app:prompts}, hyperparameters in Appendix~\ref{app:hyperparameters}, full pipeline details in Appendix~\ref{app:pipeline}, per-dataset specifications in Appendix~\ref{app:datasets}, baseline implementations in Appendix~\ref{app:baselines}, and per-method optimization-phase compute cost in Appendix~\ref{app:cost}.

\section*{Acknowledgments}

This work was supported by the Hong Kong RGC Grant RFS2425-1S01, and was done during the first author's internship at Microsoft Research Asia.

\bibliography{custom}

\appendix

\onecolumn
\nolinenumbers   
\section{Comparison with Related Work}\label{app:positioning}

\begin{table}[t]
\caption{\looseness=-1 Comparison of \method{} with prior methods.
The right block marks whether each method meets \method{}'s setting along three axes.
\textbf{Label-free}: uses no ground-truth metric or validation set.
\textbf{Full harness}: edits executable tools and skills, not memory or prompt text alone.
\textbf{Single pass}: a one-shot retrospective pass, rather than an online stream, an iterative validation-scored search, or a weight-training loop.
\cfull{}~satisfied, \chalf{}~partial, \cnone{}~not satisfied.}
\label{tab:positioning}
\centering
\footnotesize
\resizebox{\textwidth}{!}{%
\begin{tabular}{llllccc}
\toprule
& & & & \multicolumn{3}{c}{\textbf{Satisfies \method{}'s setting}} \\
\cmidrule(l){5-7}
\textbf{Method} & \textbf{Harness architecture} & \textbf{Feedback signal} & \textbf{Cost} & \makecell{\textbf{Label-}\\\textbf{free}} & \makecell{\textbf{Full}\\\textbf{harness}} & \makecell{\textbf{Single}\\\textbf{pass}} \\
\midrule
\multicolumn{7}{l}{\textit{Validation-feedback optimization}} \\
\rowcolor{rowAlt}
OPRO \citep{yang2023opro}                 & Prompt                  & Validation metric                & Iterative search    & \cnone & \cnone & \cnone \\
\rowcolor{white}
DSPy \citep{khattab2023dspy}              & Prompt $+$ demos        & Validation metric                & Iterative search    & \cnone & \cnone & \cnone \\
\rowcolor{rowAlt}
TextGrad \citep{yuksekgonul2024textgrad}  & Prompt                  & Textual gradient                 & Iterative search    & \cnone & \cnone & \cnone \\
\rowcolor{white}
GEPA \citep{agrawal2025gepa}              & Prompt                  & Val.\ metric $+$ reflection      & Iterative (genetic) & \cnone & \cnone & \cnone \\
\rowcolor{rowAlt}
ADAS \citep{hu2024adas}                   & Full agent code         & Validation accuracy              & 25--30 iter.        & \cnone & \cfull & \cnone \\
\rowcolor{white}
Meta-Harness \citep{lee2026metaharness}   & Full harness code       & Search-set score                 & $\sim$20 iter.      & \cnone & \cfull & \cnone \\
\addlinespace
\multicolumn{7}{l}{\textit{Experience-based self-improvement}} \\
\rowcolor{rowAlt}
Dynamic Cheatsheet \citep{suzgun2025dynamiccheatsheet} & Cheatsheet text & Self-judgment             & Online stream       & \cfull & \chalf & \cnone \\
\rowcolor{white}
ReasoningBank \citep{ouyang2026reasoningbank} & Memory items     & LLM-as-judge                     & Online stream       & \cfull & \cnone & \cnone \\
\rowcolor{rowAlt}
MemMA \citep{lin2026memma}                & Memory entries          & Synthetic probe QA               & Online (session)    & \cfull & \cnone & \cnone \\
\rowcolor{white}
Sleep-time Compute \citep{lin2025sleeptime} & Input context         & None                             & Precompute          & \cfull & \cnone & \cfull \\
\rowcolor{rowAlt}
SkillOS$^\dagger$ \citep{ouyang2026skillos} & Skill list            & RL reward$^\dagger$              & RL training         & \chalf & \chalf & \cnone \\
\midrule
\rowcolor{rhoBlue}
\method{} (ours)                          & Tools $+$ skills $+$ instr.\ & Self-preference             & Single pass         & \cfull & \cfull & \cfull \\
\bottomrule
\end{tabular}%
}

\vspace{2pt}
{\footnotesize $^\dagger$SkillOS's reinforcement-learning reward is an LLM-as-judge self-signal rather than gold labels, but it still requires a large training corpus of task streams, a separate judge model, and weight updates.}
\end{table}

Table~\ref{tab:positioning} compares \method{} with the prior methods discussed in Section~\ref{sec:related}.
For each method we report the harness surface it edits, the feedback signal that drives it, and the cost regime it incurs, and we mark whether it meets \method{}'s setting on three criteria, namely whether optimization is label-free, whether it edits the full harness, and whether it runs as a single offline retrospective pass.
Validation-feedback optimizers reach the full-harness end of the surface axis, where ADAS and Meta-Harness rewrite executable code, but every member of this family steers the search with a ground-truth metric and iterates against it.
Experience-based methods drop the labels, yet they only curate memory or skill text online and leave the executable harness untouched, while Sleep-time Compute precomputes context offline but optimizes nothing.
As a result, each prior method satisfies at most two of the three axes, and \method{} is the only entry that satisfies all three at once.

\section{Prompts}\label{app:prompts}

This appendix collects the prompts that instantiate the five agent
operators of \method{} (\S\ref{sec:setting}, Algorithm~\ref{alg:rho},
Figure~\ref{fig:rho-pipeline}), namely $\mathrm{solve}$, the difficulty judge
used in Coreset Selection, the diagnosis analysis, $\mathrm{optimize}$,
and $\mathrm{rank}$. The prompts are reproduced verbatim. Placeholders
of the form \texttt{\{name\}} are filled at call time with the values
described under each block.

\subsection{Solve}

Every $\mathrm{solve}(h, t)$ call materializes the harness and the
task into a fresh workspace at \texttt{harness/} and \texttt{task/} and
hands the agent the wrapper instructions below. The wrapper is
task-agnostic, in that the harness directory carries all task-shaping
guidance, and the agent is told only how to read the workspace and how
to deliver a final answer. The same wrapper is reused for the
baseline rollout and for every candidate-harness rollout so that
$\mathrm{solve}$ is the only varying input.

\begin{promptbox}[lst:solve]{The $\mathrm{solve}$ wrapper prompt.}
Solve the task defined in task/prompt.md, using the information and tools available under harness/.

The harness (harness/) is a toolkit of resources and guidance that helps the agent solve tasks. It can contain any type of file - helper scripts, artifacts, environment setup, documentation with relevant context, and workflows to follow.

Workspace layout:
  harness/   - read and invoke anything here, but do not modify it
  task/      - files for this task, including prompt.md

Steps:
1. Familiarize yourself with the information and available tools in harness/.
2. Read and analyze task/prompt.md.
3. Complete the task. For code repair tasks, modify files directly under task/repo/.
4. Present your final answer in your last message, in the format prompt.md specifies (or plain prose if unspecified).
\end{promptbox}

\subsection{Coreset Selection (Difficulty Judge)}

The difficulty judge produces the score $r_i \in [0, 10]$ and the
abstract fingerprint $\phi_i$ that drive the DPP coreset selector of
\S 4.1. It sees the task description together with a length-bounded
digest of one short prior trajectory under the current harness. The
digest is truncated head/tail to a fixed token budget, and any
commands that read the task's expected-answer files are scrubbed
before the digest is shown to the judge. The judge is asked to keep
the fingerprint in task-agnostic structural vocabulary so that
fingerprints from different codebases remain comparable under cosine
similarity.

\begin{promptbox}[lst:judge]{The difficulty judge prompt used in Coreset Selection.}
Rate the difficulty of the following software engineering task and write
an abstract structural fingerprint of it. You may also use the observed
agent run below to inform your judgment.

Output a JSON object (no markdown fences, no extra text):
{
  "difficulty": <float in [0.0, 10.0]>,
  "abstract_fingerprint": "<see guide below>"
}

Difficulty scale:
- 0-2: trivial (obvious one-line fix, cosmetic change).
- 3-5: moderate (localized change, well-defined spec).
- 6-8: hard (multi-file, non-obvious design, subtle bugs).
- 9-10: very hard (cross-cutting refactor, deep reasoning required).

Abstract fingerprint guide - 3-5 sentences (~60-120 words) describing
the *shape* of the problem in vocabulary that would apply equally to
any software project. Cover:
- Failure mode: what typically goes wrong (partial propagation,
  missed boundary case, silent precedence regression, broken invariant
  under a new branch, etc.).
- Source of difficulty: what makes it hard or easy (scattered
  invariants, ambiguous spec, tight coupling, purely localized
  arithmetic, etc.).
- Technical complexity: scope (single-function / multi-file /
  cross-module / architectural), reasoning depth (local / contextual /
  global invariant tracking), type of change (bug fix / feature /
  refactor / rollback).

Do NOT mention: repository, product, company, framework, or library
names; file paths; function, class, config, or variable names;
domain-specific nouns tied to a particular codebase. Use only abstract
structural programming vocabulary (invariant, precedence, boundary,
state reconciliation, propagation, ordering, contract, etc.).

The observed agent run contains concrete file paths, library names, and
tool output. Abstract these out the same way - fingerprints describe
shapes, not the specific codebase the agent happened to run in.

The observed run is a single noisy sample, not ground truth. Do not
lower difficulty just because the agent appeared to succeed in one
attempt, and do not raise difficulty just because one attempt thrashed
on bootstrap issues. Treat the trajectory as evidence that adjusts your
prior on task difficulty and failure mode; weight it relative to what
the task description itself implies.

Example of a well-abstracted fingerprint:
  "A multi-file refactor whose difficulty comes from keeping a single
  shared invariant consistent across several independently-evolving
  modules; the typical failure mode is partial propagation, where one
  call site adopts the new contract while another silently keeps the
  old, producing a latent bug that only surfaces under a specific
  input ordering. Spec ambiguity is low but reasoning must be global -
  the change is mechanically small per site but requires tracing a
  contract through the call graph."

Task:
---
{query}
---

Observed agent run (under the current harness):
---
{trajectory_digest}
---
\end{promptbox}

\noindent\texttt{\{query\}} is the natural-language task description.
\texttt{\{trajectory\_digest\}} is the scrubbed, head/tail-truncated
digest of one prior trajectory for the same task. This difficulty
value is the score $r_i \in [0,10]$ that enters the DPP
kernel, where it is normalized by $r_i/10$ as in \S 4.1, and the
fingerprint is embedded to a unit vector $x_i$ that defines the
similarity matrix $S = X X^\top$.

\subsection{Diagnosis}

The diagnosis prompt implements
$I_t = \mathrm{rank}_{\mathrm{val}}(t, \{\tau_g\}_{g=1}^{G}) \cup \mathrm{rank}_{\mathrm{con}}(t, \{\tau_g\}_{g=1}^{G})$ (\S 4.2). The
workspace presents the agent with the original task, the shared
harness used by every rollout, and $G$ rollout directories. The agent
executes a five-step workflow, comprising per-trajectory inspection, failure-mode
analysis (\emph{self-validation}), cross-trajectory disagreement
analysis (\emph{self-consistency}), a single high-level harness
improvement direction, and a severity score that doubles as a soft
attention weight downstream. The structured JSON output binds each
field to a fixed slot so $\mathrm{optimize}$ can attend by severity.

\begin{promptbox}[lst:diagnose]{The diagnosis prompt.}
Analyze three solve trajectories for the same task.

The harness (harness/) is a toolkit of resources and guidance that helps the agent solve tasks. It can contain any type of file - helper scripts, artifacts, environment setup, documentation with relevant context, and workflows to follow.

Workspace layout:
  task/              - the original task. Read task/prompt.md to understand the question.
  harness/           - the shared harness used by all three trajectories.
  trajectory_0/      - first solve attempt. Contains events.jsonl, final_message.txt, and workspace_diff/.
  trajectory_1/      - second solve attempt. Contains events.jsonl, final_message.txt, and workspace_diff/.
  trajectory_2/      - third solve attempt. Contains events.jsonl, final_message.txt, and workspace_diff/.

Your job is to analyze and evaluate the three trajectories. Follow this workflow:

## Step 1: Inspect each trajectory

For each of trajectory_0, trajectory_1, and trajectory_2:

1. Inspect final_message.txt and events.jsonl to understand the action and decision process.
2. Evaluate whether the trajectory accurately and efficiently completed the task.
3. Set successful to 1 if the trajectory accurately completed the task, otherwise set it to 0.
4. In quality_analysis, note what evidence, files, tools, or reasoning steps the trajectory relied on, and whether there was unnecessary work, missed information, misleading evidence, or an incorrect decision.

## Step 2: Analyze failure modes

If all three trajectories accurately and efficiently completed the task, this section can be brief. Otherwise, analyze why one or more trajectories failed or performed poorly. Make this analysis faithful and actionable. Ground it in what the trajectories actually did.

## Step 3: Analyze inconsistency

Compare the three event sequences and final answers. Identify whether there are inconsistencies among them: where and why the trajectories diverged, and how those differences affected the behavior.

## Step 4: Summarize harness improvement direction

Suggest one high-level, simple direction for improving the harness. This should be a general improvement direction based on the trajectory analysis, not a detailed edit plan and not a task-specific hardcoded fix.

## Step 5: Assign severity

Set severity to a float from 0.0 to 1.0 for how strongly this task should influence the next harness optimization:

- 0.0: no meaningful issue; all trajectories answered accurately and efficiently.
- 0.1-0.3: minor inefficiency or weak concern; do not optimize from this alone.
- 0.4-0.7: mixed success, inconsistency, or a plausible harness gap.
- 0.8-1.0: clear failure, missing information, or a high-confidence harness issue.

## Output format

Your final message must be exactly one JSON object (no markdown fences, no other text):
{
  "task_id": "<task id, if available from the task prompt or context>",
  "severity": 0.0,
  "trajectory_analyses": [
    {
      "trajectory": "trajectory_0",
      "successful": 1,
      "quality_analysis": "<faithful analysis of whether this trajectory completed the task accurately and efficiently>",
      "issues": "<any missed information, misleading evidence, inefficiency, or incorrect decision; empty string if none>"
    },
    {
      "trajectory": "trajectory_1",
      "successful": 1,
      "quality_analysis": "<faithful analysis of whether this trajectory completed the task accurately and efficiently>",
      "issues": "<any missed information, misleading evidence, inefficiency, or incorrect decision; empty string if none>"
    },
    {
      "trajectory": "trajectory_2",
      "successful": 1,
      "quality_analysis": "<faithful analysis of whether this trajectory completed the task accurately and efficiently>",
      "issues": "<any missed information, misleading evidence, inefficiency, or incorrect decision; empty string if none>"
    }
  ],
  "failure_mode_analysis": "<actionable analysis in markdown of why any trajectory failed or performed poorly; brief if none failed>",
  "inconsistency_analysis": "<root-cause analysis in markdown of where and why the trajectories diverged, and how that affected quality>",
  "harness_improvement_direction": "<one high-level, simple direction for improving the harness>"
}
\end{promptbox}

\noindent\looseness=-1 The trajectory directories carry the full event stream, the
agent's final message, and any workspace diff produced by the rollout.
The \texttt{$-$self-validation} and \texttt{$-$self-consistency}
ablations in Table~\ref{tab:diagnosis} remove the corresponding step
and the corresponding output field, and the surrounding scaffolding is
left intact.

\subsection{Optimization}

The optimization prompt implements
$h_j = \mathrm{optimize}(h_0, \{I_t\}_{t \in \mathcal{D}_{\mathrm{core}}})$
(\S 4.3). The editor agent is given write access to a fresh copy of
the harness and a directory of per-task diagnoses sorted by severity.
The instruction frames severity as a soft attention weight, requires
cross-task pattern matching before any edit, and discourages
task-specific hardcoded fixes. Each of the $N$ candidates is sampled
independently with the same prompt, and randomness from the underlying
agent supplies the diversity.

\begin{promptbox}[lst:optimize]{The $\mathrm{optimize}$ prompt.}
Based on the per-task diagnoses in diagnoses/, analyze and optimize the current harness/ to improve performance on future tasks. "Better performance" means the agent's final answer more directly and correctly answers what each task asks, with fewer wasted steps.

The harness (harness/) is a toolkit of resources and guidance that helps the agent solve tasks. It can contain any type of file - helper scripts, artifacts, environment setup, documentation with relevant context, and workflows to follow.

Workspace layout:
  harness/       - the current harness; you may directly modify it (add/remove/edit any files)
  diagnoses/     - one subdirectory per task, each containing:
    - diagnosis.md  - structured trajectory analysis, severity, failure modes, inconsistency analysis, and a high-level harness improvement direction
    - prompt.md     - the original task for context

## Steps

1. Read each diagnosis in diagnoses/task_XXXX/diagnosis.md and the corresponding prompt.md.
2. Use Severity as a soft attention weight from 0.00 to 1.00, not as ground truth. Higher severity means the diagnosis should influence optimization more strongly.
3. Look for patterns across diagnoses - are multiple tasks failing for similar reasons?
4. Prioritize high-severity recurring failure modes and high-severity recurring inconsistency root causes.
5. Low-severity tasks usually should not cause a harness edit by themselves unless the same issue motif recurs across tasks.
6. Make surgical improvements to harness/ that address the high-level diagnosed issues.

When done, send the changes and your rationale as your final message. If you made no changes, explain why the current harness is already sufficient.
\end{promptbox}

\noindent Each \texttt{diagnoses/task\_XXXX/} subdirectory holds the
serialized diagnosis JSON rendered as Markdown alongside the original
task prompt. Subdirectories are indexed in descending severity so the
agent encounters the most consequential diagnoses first.

\subsection{Pairwise Ranking}

\looseness=-1 The ranking prompt implements
$\mathrm{rank}(t, \tau_a, \tau_b) \in [-10, 10]$ used in Best-of-$N$
acceptance (\S 4.3 and Algorithm~\ref{alg:rho}). The evaluator sees the task, the
two harnesses, and the two trajectories. The candidate trajectory is
presented as \texttt{trajectory\_A} and the baseline as
\texttt{trajectory\_B}, and the orchestrator negates the returned integer
so the scalar score is oriented as baseline\,$\to$\,candidate
regardless of presentation order. Presenting the candidate first
reduces a later-option preference bias we observed in pilot runs.
The rubric below uses an integer scale in $[-10,10]$, and downstream only
the sign and relative magnitude of the score are used (\S\ref{sec:setting}).

\begin{promptbox}[lst:rank]{The $\mathrm{rank}$ prompt used in Best-of-$N$ acceptance.}
Analyze the difference in performance between harness A and harness B on the same task. You will see one task/ directory (containing prompt.md) and two trajectory_*/ directories (each containing events.jsonl and final_message.txt). Produce a JSON object scoring the A -> B transition on an integer scale from -10 to +10 with a short rationale.

The harness (harness/) is a toolkit of resources and guidance that helps the agent solve tasks. It can contain any type of file - helper scripts, artifacts, environment setup, documentation with relevant context, and workflows to follow.

Scoring rubric:

- +10: A -> B is a change from unacceptable to excellent; B's trajectory is efficient and its answer is correct.
- 0: A and B perform comparably, or it is not possible to determine which is better.
- -10: A -> B is a severe regression; B's trajectory is inefficient and its answer is wrong.

Workspace layout:
  task/                 - the original task (prompt.md is the question)
  harness_A/            - the harness used by trajectory A
  harness_B/            - the harness used by trajectory B
  trajectory_A/         - trajectory from harness A (final_message.txt is the answer)
  trajectory_B/         - trajectory from harness B (final_message.txt is the answer)

## Evaluation steps

1. Read task/prompt.md to understand what the task requires.
2. Read and compare trajectory_A and trajectory_B.

Your final reply must be exactly one JSON object (no markdown fences, no other text):
{"value": <integer in [-10, 10]>, "rationale": "<one-sentence rationale>"}
\end{promptbox}

\noindent The candidate score $S_j$ in Algorithm~\ref{alg:rho} averages this
integer (with the orientation flip described above) over the coreset
$\mathcal{D}_{\mathrm{core}}$, and the candidate is accepted only when
$S_j > 0$, otherwise the harness remains at $h_0$.

\linenumbers
\twocolumn
\section{Hyperparameters and Infrastructure}\label{app:hyperparameters}

Table~\ref{tab:hparams} lists every hyperparameter and infrastructure setting used in our experiments.
Values already reported in the main text (Section~\ref{sec:setting} and Section~5.1) are reproduced here for convenience, and values not previously stated are introduced here.
The intent is an audit-ready specification, with one column per parameter, one value per cell, and no defaults left implicit.
Dataset-specific overrides on solver wall-clock and grading are deferred to Appendix~\ref{app:datasets}, and prompt text for every agent role is in Appendix~\ref{app:prompts}.

\begin{table*}[t]
\caption{Hyperparameters and infrastructure for \method{}. Parameters appearing in the main text are marked with a $^{\dagger}$ and are restated here for completeness only.}
\label{tab:hparams}
\centering
\small
\begin{tabular}{lll}
\toprule
\textbf{Parameter} & \textbf{Value} & \textbf{Description} \\
\midrule
\multicolumn{3}{l}{\textit{Backbone agent}} \\
\midrule
Model$^{\dagger}$           & Codex \texttt{gpt-5.5}                       & Shared across solve, optimize, and rank. \\
Reasoning effort$^{\dagger}$ & high                                        & Applied to all roles uniformly. \\
Sampling temperature        & provider default                             & Codex reasoning models do not expose temperature. \\
Provider                    & Cloud-hosted \texttt{gpt-5.5} (Codex CLI)    & Single hosted endpoint. \\
\midrule
\multicolumn{3}{l}{\textit{Coreset selection}} \\
\midrule
Selector                    & DPP, greedy MAP                              & \cite{kulesza2012dpp}. \\
Coreset size $k$$^{\dagger}$ & 10                                          & Train tasks. \\
DPP weight $\theta$$^{\dagger}$ & 0.7                                      & Difficulty/diversity tradeoff. \\
Score floor $\epsilon$      & 0.1                                          & Lower bound on normalized difficulty. \\
Judge model                 & Codex \texttt{gpt-5.5}, high                 & Same backbone as the solver. \\
Trajectory-digest budget    & 10{,}000 BPE tokens, head/tail truncation    & Ground-truth-revealing commands scrubbed. \\
Fingerprint embedding       & \texttt{BAAI/bge-large-en-v1.5}, 1024-d      & Executed locally \citep{xiao2023cpack}. \\
\midrule
\multicolumn{3}{l}{\textit{Group rollout}} \\
\midrule
Rollouts per task $G$       & 3                                            & Parallel solves under $h_0$ per coreset task. \\
Solve wall-clock timeout    & 900 s                                        & Dataset overrides in Appendix~\ref{app:datasets}. \\
Diagnosis prompt            & see Appendix~\ref{app:prompts}               & Self-validation $+$ self-consistency cues. \\
Severity range              & $[0,1]$                                      & Soft attention weight, not a hard threshold. \\
\midrule
\multicolumn{3}{l}{\textit{Best-of-$N$ harness proposal}} \\
\midrule
Candidates $N$$^{\dagger}$  & 3                                            & Parallel optimizer samples. \\
Optimizer prompt            & see Appendix~\ref{app:prompts}               & Conditioned on $\{I_t\}_{t\in\mathcal{D}_{\mathrm{core}}}$. \\
Ranking timeout             & 300 s per pairwise call                      & Per $\mathrm{rank}(t,\tau_a,\tau_b)$. \\
Acceptance threshold        & $S_j > 0$ (strict)                           & Mean pairwise score over the coreset. \\
\midrule
\multicolumn{3}{l}{\textit{Infrastructure}} \\
\midrule
Agent-call concurrency      & 10 concurrent calls (cap 30)                 & Parallelism for Table~\ref{tab:main} runs. \\
Rounds per experiment       & 1                                            & Unless stated otherwise. \\
Persistence                 & full logs to disk                            & Prompts, completions, trajectories, diagnoses, \\
                            &                                              & candidate harnesses, diffs, scores, run summaries. \\
\bottomrule
\end{tabular}
\end{table*}

Several choices in Table~\ref{tab:hparams} warrant brief justification.
We set $G=3$ because three rollouts are enough for the diagnosis prompt to surface cross-trajectory disagreements, and pilots with larger $G$ inflated cost linearly without sharpening the diagnosis signal.
We weight the DPP at $\theta=0.7$ so that difficulty dominates diversity, on the principle that an unselected easy task carries less optimization signal than a redundant hard one.
Concretely, $\theta$ enters through the difficulty weights $\widetilde{r}_i = \big(\max(r_i,\epsilon)\,/\,\max_j \max(r_j,\epsilon)\big)^{\alpha}$ with $\alpha = \theta/\big(2(1-\theta)\big)$, and the factor of two offsets the difficulty term's doubled appearance in $\log\det K$, so that $\theta$ and $1-\theta$ weight the difficulty and diversity terms directly.
We make the acceptance gate strictly positive rather than non-negative because pairwise self-preference is a noisy estimator, and breaking ties in favor of change would inflate regression risk for no expected gain.
Finally, we use the same Codex \texttt{gpt-5.5} backbone for solver, optimizer, and ranker, since decoupling the solver from the judge would introduce a confound where measured held-out gains could be attributed to a stronger judge rather than to \method{} itself.
Per-dataset configuration overrides referenced above are in Appendix~\ref{app:datasets}, and baseline-specific configurations are in Appendix~\ref{app:baselines}.

\section{Pipeline Implementation Details}\label{app:pipeline}

\looseness=-1 This appendix documents the design choices that govern \method{}'s behavior but do not appear in \S 4 or Algorithm~\ref{alg:rho}. The equations and operator signatures are fixed by the main text, and what follows are the mechanics that make them executable.

\subsection{Harness Representation and Mounting}

\looseness=-1 A harness is a directory of files with no fixed schema, where prose instructions, executable scripts, and structured configuration sit side by side. At every operator call the harness is materialized as a subdirectory of the agent's working directory, and the agent reads it the same way it reads task files rather than as a system-prompt prefix. The $\mathrm{solve}$ operator mounts the harness read-only by convention, while $\mathrm{optimize}$ mounts a fresh copy with write access and accepts whatever the agent leaves behind on the filesystem. Harnesses are compared by content, so if $\mathrm{optimize}$ returns a harness identical to the input, the candidate is treated as a no-op and dropped before evaluation. This keeps the harness surface tool-agnostic and lets the optimizer extend any of the three file kinds without a representation change.

\subsection{Role Separation and Workspace Isolation}

\looseness=-1 The same backbone executes $\mathrm{solve}$, the diagnosis analysis ($\mathrm{rank}_{\mathrm{val}}$/$\mathrm{rank}_{\mathrm{con}}$), $\mathrm{optimize}$, and $\mathrm{rank}$. Role separation is achieved by workspace contents rather than by changing the backbone, where each operator runs in a fresh workspace that contains only the inputs it should see. $\mathrm{solve}$ sees the task files and the current harness. The diagnosis analysis sees the task and the group of trajectories for that task. $\mathrm{optimize}$ sees the harness directory, the diagnosis instructions, and the trajectory rollouts that motivated them. Finally, $\mathrm{rank}$ sees the task, two trajectories, and both candidate harness directories side by side. Per-role ablation therefore amounts to swapping prompts and inputs, not models, which keeps the operators directly comparable.

\subsection{Order, Parsing, and Failure Handling in Pairwise Ranking}

\looseness=-1 $\mathrm{rank}$ is invoked once per (task, candidate) pair. We present the candidate trajectory first and the baseline trajectory second, then negate the parsed scalar. This swap is a standard mitigation for position bias in pairwise judges, applied without per-call order randomization. The judge is constrained to return a single integer in $[-10, 10]$ with a one-sentence rationale, and we do not retry. Any parse or execution failure deterministically yields zero. This makes $\mathrm{rank}$ a strict pessimist, since silent failures pull the mean toward the rejection threshold rather than away from it.

\subsection{Diagnosis vs.\ Ranking Inputs}

The diagnosis analysis consumes the full group of $G$ rollouts for one task and produces a single textual instruction $I$ with a severity weight, whereas $\mathrm{rank}$ is strictly pairwise. We do not collapse a group into a Bradley--Terry score for evaluation. For each candidate, we pair the candidate's rollout of a task against a fixed baseline trajectory drawn from the original group, and we hold this baseline constant across every candidate so the comparison stays anchored to the same reference. The consequence is that $G > 1$ enters the pipeline as a diagnostic device only, since it sharpens the instruction $I$ through cross-trajectory inconsistency but does not vote on which candidate wins.

\subsection{Optimizer Action Space}

$\mathrm{optimize}$ is a code-agent invocation, not a constrained text rewriter. It sees the materialized harness as a filesystem and may add, remove, or modify any file inside it. We hand it the diagnosis instructions sorted by descending severity, with severity exposed as a soft attention weight in $[0, 1]$ rather than a hard priority queue, so the agent can still merge or skip suggestions that conflict. We do not parse a diff out of the agent's final message, and the new harness state is recovered from the directory itself when the call returns. This keeps the action space identical to the representation, where anything that fits in a directory is a valid edit.

\subsection{Acceptance Gate and No-Ops}

An update is accepted only when the best candidate's mean pairwise score over the coreset is strictly positive, and a mean of zero is rejected. Before the gate, a candidate is dropped if the optimizer failed, the call timed out, or the produced harness is identical to the input. If every candidate is dropped, or every surviving candidate scores $S_j \leq 0$, \method{} leaves the harness unchanged. This no-update behavior is part of the result rather than a tooling failure. In Figure~\ref{fig:coreset-schematic}, selectors whose coreset fails to expose useful weaknesses (pure coverage, pure difficulty) leave held-out accuracy near the Vanilla Codex baseline.

\subsection{Persistence}

We persist the input harness id, along with the trajectories produced by each operator ($\mathrm{solve}$, the diagnosis analysis, $\mathrm{optimize}$, and $\mathrm{rank}$) identified by id. We further persist the diagnosis instruction, the candidate harness ids together with their diffs against the input, the per-candidate pairwise scores, the mean pairwise score, and the accept flag. Every stored trajectory carries its full event stream, final message, workspace diff, and wall-clock time. These records are what makes downstream audit, re-grading, and ablation possible without re-running the agent.

\section{Dataset Specifications}\label{app:datasets}

\looseness=-1 This appendix documents what we take from each upstream benchmark, how we split it, what the agent sees at solve time, and how grading runs.
Pinned upstream commits make the partition and grader reproducible from a clean checkout.

\subsection{SWE-Bench Pro}

SWE-Bench Pro \citep{deng2025swebenchpro} is the long-horizon software-engineering benchmark in our suite.
Tasks resolve only when a multi-file patch passes the upstream test set, and failure modes are therefore concrete and traceable to the harness's repository conventions and build commands.

\noindent\textbf{Source.}
We load the test split of \texttt{ScaleAI/SWE-bench\_Pro} from Hugging Face.
Evaluator scripts and per-instance Docker images come from the official \texttt{scaleapi/SWE-bench\_Pro-os} repository, pinned to a fixed commit.\footnote{Commit \texttt{0c64e26f00b9}.}

\noindent\textbf{Split.}
Rows are ordered by the SHA-256 hash of \texttt{(seed, instance\_id)} with the seed fixed.
The first 100 form the training pool and the next 100 the held-out test pool, and the remaining rows are unused.
We report on the held-out test pool.

\looseness=-1\noindent\textbf{Solve interface.}
Each task materializes a prompt file describing the issue plus a fresh clone of the upstream repository at its base commit, mounted in the workspace.
The agent edits repository files in place, and no extra tools are injected.

\noindent\textbf{Grading.}
We extract the agent's patch by re-applying its workspace edits to a fresh checkout and taking \texttt{git diff --binary}.
If the workspace path fails, we fall back to scanning the agent's final message for fenced diff blocks.
Binary hunks and auto-generated paths (dependency, cache, and build directories) are stripped before scoring.
The patch is then applied inside the official per-instance Docker image distributed by the SWE-Bench Pro authors, and the official run and parser scripts are invoked.
A task passes iff every \texttt{FAIL\_TO\_PASS} and \texttt{PASS\_TO\_PASS} test resolves correctly.
The Docker wall-clock budget is one hour per task.

\subsection{Terminal-Bench~2}

Terminal-Bench~2 \citep{terminalbench2} contains executable command-line tasks. Failure is a missed reward, not a missed convention, so the harness mostly buys consistency in how the agent inspects state and chains commands.

\noindent\textbf{Source.}
We use the upstream Terminal-Bench~2 repository at a fixed commit\footnote{Commit \texttt{53ff2b87d621}.}, yielding 89 tasks.

\noindent\textbf{Split.}
Tasks are hash-ordered with the same seeded scheme as SWE-Bench Pro.
The first 30 form the training pool, the remaining 59 the held-out pool.

\noindent\textbf{Solve interface.}
Each task spins up a fresh Docker container whose image is declared by the task.
Containers carry a cleanup label and per-task CPU and memory limits.
The agent's host workspace is bind-mounted inside the container.
The solve prompt instructs the agent to author shell scripts on the host and execute them inside the container.
A wall-clock watchdog terminates the container when the task's declared agent timeout elapses.

\looseness=-1\noindent\textbf{Grading.}
We run the task's upstream test suite inside the still-running container under its declared verifier timeout.
The verifier writes a reward of 0 or 1 to a known path.
A task passes iff the reward equals 1.
We apply no difficulty filter, so held-out numbers span the upstream difficulty mix.

\noindent\textbf{Data integrity.}
The solve prompt instructs the agent not to read the test directory, the upstream solution, or the verifier log.
This is a prompt-level convention, not a sandbox restriction.
We document it for transparency, since a sufficiently adversarial agent could read the verifier and game the reward.
We rely on the held-out pool and Table~\ref{tab:main}'s cross-benchmark consistency to detect such behavior, and we have not observed it.

\subsection{GAIA-2}

\looseness=-1 GAIA-2 \citep{froger2026gaia2} differs from the two coding benchmarks in that the environment evolves independently of the agent.
The harness layer therefore has to encode behavior under partial observation and asynchronous events, not just stable build conventions.

\noindent\textbf{Source.}
We load the validation split of the GAIA-2 release on Hugging Face under the \texttt{mini} configuration, yielding 200 scenarios.
Each scenario ships an asynchronous event stream and an upstream write-action verifier.

\noindent\textbf{Split.}
Scenarios are hash-ordered with the same seeded scheme.
The first 100 form the training pool, and the remainder, 100, form the held-out pool.
We report on the held-out slice.

\looseness=-1\noindent\textbf{Solve interface.}
Each scenario's workspace contains a task prompt, a tool dispatcher, and a tool catalog.
The agent invokes tools through a single shell command per call.
The dispatcher relays each call to a sidecar process running the upstream environment, which advances simulated time and replays scheduled events.
The scenario's ground-truth state is held entirely by the sidecar, and the agent never reads it.

\looseness=-1\noindent\textbf{Grading.}
When the agent finishes, the sidecar invokes the scenario's upstream verifier.
The judge LLM is the same Codex \texttt{gpt-5.5} backbone we use elsewhere, and routing defaults to the same Azure Foundry endpoint.
A scenario passes iff the upstream verifier reports success.

\noindent\textbf{Environment modifications.}
Two changes to the upstream environment affect reported numbers and must be disclosed.
First, the upstream cap of one \texttt{send\_message\_to\_user} call per turn is raised to four, since the original cap penalizes verbose agents that would otherwise complete the task correctly.
Second, three optional judge-relaxation switches (event filtering, relaxed per-app UI judging, and trivial filesystem-read filtering) are available as ablation axes but are all disabled in the experiments reported in Table~\ref{tab:main}.

\section{Baseline Implementations}\label{app:baselines}

\looseness=-1 Every baseline runs under the same Codex \texttt{gpt-5.5} backbone at high reasoning effort.
The trajectory-only family (Dynamic Cheatsheet, ReasoningBank, and Sleep-time Compute) shares \method{}'s Coreset Selection budget of 10 training tasks and 3 candidates.
Baselines differ only in what the offline phase persists and how the solver consumes it.
This appendix documents each.

\subsection{Dynamic Cheatsheet \citep{suzgun2025dynamiccheatsheet}}

Dynamic Cheatsheet curates a running list of reusable facts and procedures harvested from past trajectories.

\noindent\textbf{Persistence layer.}
A single markdown file lives inside the harness, structured as \texttt{<memory\_item>} blocks containing a description, a worked example, and a usage count.

\noindent\textbf{Offline phase.}
A curator agent iterates over the selected training tasks in order.
For each task it reads the current cheatsheet plus the solve transcript and rewrites the cheatsheet in place.

\noindent\textbf{Online consumption.}
The cheatsheet is part of the harness, and the solver reads it as part of normal harness lookup.
No additional prompt prefix is injected.

\noindent\textbf{Faithfulness.}
Upstream Dynamic Cheatsheet updates the cheatsheet after every task, so the next task's solver sees the update.
Our default configuration runs solves for all selected training tasks before invoking the curator, then commits one cheatsheet.
The single-task configuration that matches the upstream per-example update is available as a sensitivity setting, while Table~\ref{tab:main} uses the default.

\subsection{ReasoningBank \citep{ouyang2026reasoningbank}}

ReasoningBank stores reusable reasoning patterns extracted from past trajectories and retrieves them on demand at inference time.

\noindent\textbf{Persistence layer.}
A JSONL bank of items, each with title, description, body, and a precomputed embedding.
The bank lives outside the harness directory because solve-time access is by similarity retrieval, not by harness materialization.

\noindent\textbf{Offline phase.}
For each selected training task we solve it, ask the judge whether the trajectory was successful, then run an extraction prompt that distills reusable reasoning patterns from the trajectory.
Extracted items are appended to the bank, and embeddings are computed locally.

\noindent\textbf{Online consumption.}
At each held-out solve, we encode the task description with the same encoder, retrieve the top-$n$ items by cosine similarity, render them as a memory preamble prepended to the solve instructions.
Top-$n$ is fixed at 1.

\looseness=-1\noindent\textbf{Faithfulness.}
We replace the upstream Gemini embedding with \texttt{BAAI/bge-large-en-v1.5} \citep{xiao2023cpack}, a 1024-dimensional locally executed sentence encoder, so that every baseline that uses retrieval shares the same encoder, dimensionality, and storage backend.
This removes a confound between memory-bank quality and embedding-provider quality.

\subsection{Sleep-time Compute \citep{lin2025sleeptime}}

Sleep-time Compute preprocesses past traces into compact notes that are prepended to the agent's context at inference time, following the Letta implementation \citep{packer2023memgpt}.

\noindent\textbf{Persistence layer.}
A set of bounded markdown memory blocks stored inside the harness, edited with structured edit tools (insert, replace, rethink, finish) that mirror upstream Letta semantics.

\noindent\textbf{Offline phase.}
A sleep-time agent iterates over the selected training tasks.
For each task it reads the task's prompt and trajectories, then issues a sequence of memory-edit tool calls to update the harness's memory blocks.
The system prompt is the upstream Letta system prompt, used verbatim.

\noindent\textbf{Online consumption.}
The solver reads the memory blocks as part of the harness, the same way it reads any harness file.

\looseness=-1\noindent\textbf{Scope.}
Edits accumulate across tasks, and the sleep-time agent works against a single evolving memory state task after task.
We do not reset memory between training tasks.

\subsection{Meta-Harness \citep{lee2026metaharness}}

Meta-Harness is the validation-feedback reference, where a meta-agent rewrites the harness, and the rewrites are scored against a labeled validation set.

\noindent\textbf{Persistence layer.}
A sequence of candidate harness directories paired with a search history recording per-candidate validation pass rates.

\noindent\textbf{Offline phase.}
An outer loop alternates between a proposer that reads the search history and emits a new harness, and an evaluator that grades the proposed harness on the search-task set using the dataset's ground-truth grader.
Our re-implementation's default budget is 20 outer iterations $\times$ 3 candidates per iteration $\times$ 2 solve trials per task, far above the budget we give \method{}; upstream reports evaluating roughly 60 candidate harnesses over 20 iterations.

\looseness=-1\noindent\textbf{Validation-feedback footprint.}
The proposer reads each candidate's per-task scores, mean score, and pass rate from the history, and the grader signal directly shapes the next proposal.
This is the hallmark of the validation-feedback family and the axis along which Meta-Harness is incomparable to the trajectory-only baselines.

\noindent\textbf{Matched-budget configuration.}
For Table~\ref{tab:meta-harness} we run the outer loop for one iteration with 3 candidates and 1 solve trial per task, so the candidate count matches \method{}'s $N=3$.
The validation-grade cost is what Table~\ref{tab:meta-harness} reports.

\looseness=-1\noindent\textbf{Shared configuration.}
All four baselines and \method{} start from the same empty harness for a given dataset, use the same coreset of selected training tasks, and are graded against the same held-out split with the same grader.
The only deliberate axis of variation is what the offline phase persists and how the solver consumes it.
Meta-Harness additionally consumes validation grades, and this is the axis Table~\ref{tab:meta-harness} measures.

\begin{table*}[t]
\caption{Agent invocations by role for the optimization phase on SWE-Bench Pro ($k=10$ train tasks, $M_{\mathrm{test}}=100$ held-out tasks). \textsc{rollout} is $G$ parallel solves per coreset task, \textsc{after} is one solve per candidate per coreset task, \textsc{rank} is one pairwise rank per candidate per coreset task, and \textsc{test} is one solve per held-out task. Counts are read directly from the persisted trajectory directories, and \textsc{llm} reports auxiliary chat-completion calls.}
\label{tab:cost-breakdown}
\centering
\small
\begin{tabular}{lrrrrrrrr}
\toprule
\textbf{Method} & \textbf{rollout} & \textbf{diagnose} & \textbf{optimize} & \textbf{after} & \textbf{rank} & \textbf{test} & \textbf{total} & \textbf{llm} \\
\midrule
Vanilla Codex                              &   0 &  0 &  0 &  0 &  0 & 100 & 100 &  0 \\
ReasoningBank \citep{ouyang2026reasoningbank}   &  10 &  0 &  0 &  0 &  0 & 100 & 110 & 20 \\
Dynamic Cheatsheet \citep{suzgun2025dynamiccheatsheet} &  30 &  0 & 10 & 10 & 10 & 100 & 160 &  0 \\
Sleep-time Compute \citep{lin2025sleeptime}        &  30 &  0 & 30 & 30 & 30 & 100 & 220 &  0 \\
\method{}                            &  30 & 10 &  3 & 30 & 30 & 100 & 203 &  0 \\
\bottomrule
\end{tabular}
\end{table*}

\section{Optimization-Phase Compute Cost}\label{app:cost}

This appendix documents the optimization-phase compute cost of \method{} and the baselines reported in Table~\ref{tab:main} and Table~\ref{tab:meta-harness}.
We report the numbers without claiming an efficiency advantage. Among the trajectory-only methods at matched coreset budget ($k=10$, $N=3$), \method{} sits in the same order of magnitude as Sleep-time Compute and Dynamic Cheatsheet, and is more expensive than ReasoningBank.
The matched-budget Meta-Harness configuration is the only baseline whose cost is qualitatively different, and its upstream-default budget would be larger still, as we discuss at the end.

\subsection{Accounting Scope}

Every method dispatches two kinds of model calls during the offline phase, plus the same per-task solve at evaluation time.
We count them separately because they go through different backends and have different unit costs.

\noindent\textbf{Codex (agent) invocations.}
One Codex CLI subprocess that runs to completion is one agent invocation.
Each invocation persists a trajectory directory recording the role (one of solve, diagnose, optimize, rank), the harness id, the task id, the wall-clock duration, and the exit status.
We obtain the per-method totals by counting these directories.

\noindent\textbf{Auxiliary LLM calls.}
A single direct API call to the same \texttt{gpt-5.5} backbone, issued outside the Codex CLI.
Only ReasoningBank uses these in the offline phase, namely two per training task, for success judgment and memory-item extraction.
Coreset selection also issues one auxiliary LLM call per candidate task before any method starts. This cost is shared across methods at matched coreset and is reported once at the end of this appendix.

\looseness=-1\noindent\textbf{Embedding calls.}
ReasoningBank and DPP coreset selection both call \texttt{BAAI/bge-large-en-v1.5} \citep{xiao2023cpack} for fingerprint or query embedding.
We execute the encoder locally, so these calls do not hit a remote endpoint and are not aggregated with the LLM totals.

\subsection{Per-Method Decomposition}

Table~\ref{tab:cost-breakdown} breaks down the optimization phase on SWE-Bench Pro ($k=10$ train tasks, $M_{\mathrm{test}}=100$ held-out tasks) into agent invocations by role.
Terminal-Bench~2 differs only in $M_{\mathrm{test}}=59$, while GAIA-2 matches SWE-Bench Pro.

\looseness=-1 The table is a literal accounting of how each method spends its offline budget.
ReasoningBank issues one solve per training task and never proposes a candidate harness, so the offline phase costs only $k=10$ extra Codex invocations on top of the held-out evaluation, and its $2k$ auxiliary LLM calls are the success judge and the memory-item extractor.
Dynamic Cheatsheet and Sleep-time Compute both roll out each training task $G=3$ times to surface variance, then run an optimizer agent. Dynamic Cheatsheet invokes a single curator over all $k$ tasks ($N{=}1$ candidate), while Sleep-time Compute runs one edit pass per task per restart (3 restarts $\times$ $k$ tasks $=30$ optimize calls).
Both then commit each unique candidate harness through the after-solve and pairwise-rank steps shared with \method{}.
\method{}'s footprint differs from Sleep-time Compute's only by trading $30-3=27$ per-task optimize calls for $10$ diagnose calls plus the $3$ candidate-level optimize calls in the best-of-$N$ proposer, and the after-solve and rank phases are identical at $N=3$ candidates.

\subsection{Wall-Clock Time}

\looseness=-1 Table~\ref{tab:cost-walltime} reports two complementary wall-clock measurements per (method, dataset) cell.
$\Sigma_{\mathrm{codex}}$ is the sum of per-invocation wall-clock over every agent invocation in the run, and it is the cost of running the same workload serially on one agent process.
End-to-end is the elapsed time from the run's start to its completion, under our infrastructure setting of 10 concurrent agent calls and 10 concurrent graders.
The ratio between the two columns reflects how much of the workload was successfully parallelized, where the trajectory-only methods reach $5$--$10\times$ speedup, while ReasoningBank reaches $3.5$--$4.5\times$ because its offline phase is sequential by construction (each training task's solve must commit to memory before the next task's retrieval).
Meta-Harness is a validation-feedback rather than a trajectory-only method, so we report its budget separately in \S\ref{app:cost:mh}, not here.

\begin{table*}[t]
\caption{Wall-clock cost per run, by dataset. $\Sigma_{\mathrm{codex}}$ is the sum of per-invocation wall-clock over all agent invocations (i.e. the cost on one Codex subprocess), and \emph{end-to-end} is the elapsed time from run start to completion under 10-way agent concurrency and 10-way grader concurrency. Hours.}
\label{tab:cost-walltime}
\centering
\small
\begin{tabular}{l rr rr rr}
\toprule
& \multicolumn{2}{c}{\textbf{SWE-Bench Pro}} & \multicolumn{2}{c}{\textbf{Terminal-Bench~2}} & \multicolumn{2}{c}{\textbf{GAIA-2}} \\
\cmidrule(lr){2-3} \cmidrule(lr){4-5} \cmidrule(lr){6-7}
\textbf{Method} & $\Sigma_{\mathrm{codex}}$ & end-to-end & $\Sigma_{\mathrm{codex}}$ & end-to-end & $\Sigma_{\mathrm{codex}}$ & end-to-end \\
\midrule
Vanilla Codex                                 &  9.4 & 1.1 &  4.3 & 0.6 &  4.2 & 0.6 \\
ReasoningBank \citep{ouyang2026reasoningbank}      & 17.5 & 4.6 &  8.0 & 2.4 &  7.2 & 2.0 \\
Dynamic Cheatsheet \citep{suzgun2025dynamiccheatsheet} & 17.9 & 2.5 & 10.8 & 1.9 &  6.5 & 1.2 \\
Sleep-time Compute \citep{lin2025sleeptime}           & 22.2 & 3.2 & 14.5 & 2.2 &  7.0 & 1.2 \\
\method{}                                & 23.1 & 3.2 & 15.5 & 2.5 &  9.2 & 1.4 \\
\bottomrule
\end{tabular}
\end{table*}

Two observations are worth stating explicitly.
First, per-invocation cost varies by more than $2\times$ across datasets (GAIA-2 averages around $150$\,s per invocation, Terminal-Bench~2 around $300$\,s, SWE-Bench Pro around $400$\,s), so an agent-call count is informative only within a dataset, and cost cannot be compared across datasets by summing invocation counts alone.
Second, at matched coreset and $N=3$, \method{} is within $\pm 4\%$ of Sleep-time Compute on every dataset, more expensive than Dynamic Cheatsheet (which uses $N=1$), and substantially more expensive than ReasoningBank (which performs no group rollout, no candidate proposal, no pairwise rank).

\subsection{Matched-Budget Meta-Harness}\label{app:cost:mh}

Meta-Harness has a qualitatively larger compute footprint because validation grades feed back into the proposer.
At its upstream default ($I=20$ outer iterations, $C=3$ candidates per iteration, $T=2$ solve trials per task) the offline phase alone would cost
$I (1 + C M_{\mathrm{search}} T) + M_{\mathrm{search}} T \approx 1{,}210$
agent invocations on SWE-Bench Pro --- nearly twelve times \method{}'s optimization-phase count of $103$.
The matched-budget configuration we report in Table~\ref{tab:meta-harness} reduces this to $I=1$, $C=3$, $T=1$, amounting to $M_{\mathrm{search}} T + I (1 + C M_{\mathrm{search}} T) = 10 + 31 = 41$ optimization-phase invocations; this is the configuration whose held-out pass rate ($0.62$ on SWE-Bench Pro) is reported alongside \method{}'s in the main text.
The 10-round configuration reported in the same table keeps $C=3$, $T=1$ but raises $I$ to $10$, giving $10 + 10 \cdot 31 = 320$ optimization-phase invocations ($3.1\times$ \method{}'s optimization-phase count of $103$).
These counts exclude the shared $M_{\mathrm{test}}=100$ held-out evaluation, which every method pays once regardless of optimizer.
Because Meta-Harness is a validation-feedback rather than a trajectory-only method, it is not included in the wall-clock comparison of Table~\ref{tab:cost-walltime}.

\subsection{Coreset Selection Cost (Shared)}

\looseness=-1 Coreset selection runs once before any method's offline phase begins.
For SWE-Bench Pro this consumes $100$ auxiliary LLM calls (one difficulty judgment per candidate task) plus one batched embedding call over the same $100$ fingerprints, after which the selected $k=10$ task ids are persisted and reused verbatim by every baseline.
This cost is paid once per dataset and is not attributed to any individual method.

\onecolumn
\nolinenumbers
\section{Optimized Harness Artifacts}\label{app:artifacts}

This appendix reproduces, verbatim, the complete contents of the highest-scoring harness \method{} produced on each benchmark, i.e.\ the harnesses summarized in Figure~\ref{fig:harness-artifacts}. Each harness is a directory of Markdown instruction and skill files together with executable tool scripts; every file is shown in full, with no abridgement. Files are grouped by benchmark and labeled with their path inside the harness directory and their role (instruction, skill, or tool).

\subsection{SWE-Bench Pro}

\begin{promptbox}[lst:art-swe-readme]{SWE-Bench Pro harness: \texttt{README.md} (instructions)}
# SWE-bench Repair Harness

Use this harness before editing and before finalizing a task. The goal is to
make the final patch answer the exact prompt, not just compile.

## Default Workflow

1. Read the prompt as a contract. Write a short requirement ledger covering:
   exact public names, file paths, payload shapes, errors/status codes,
   default values, supported backends/platforms, and called-out edge cases.
2. Look for an existing reference when the task sounds like a known fix:
   local git history, tags, upstream commits, tests, fixtures, or nearby
   implementation patterns. Prefer adapting proven behavior over guessing.
3. Trace the real path end to end before editing. Follow stored data back to
   user-visible artifacts, config values back to their runtime representation,
   and wrappers back to the implementation that actually executes.
4. Make the smallest production-source change that satisfies the contract.
   Keep generated files, caches, temporary smoke tests, and local environment
   output out of the submitted diff.
5. Verify executable behavior. Syntax checks are not enough: run at least one
   focused smoke that calls every new public route/function/method and covers
   each explicit failure/edge condition in the prompt.
6. Re-open the requirement ledger before the final answer. Check literal API
   names, response shapes, defaults, cleanup behavior, and every backend or
   platform named by the prompt.

## Tools

- `harness/bin/repair-verify` discovers common non-PATH toolchains, runs cheap
  language checks, reports generated artifacts, and accepts extra task-specific
  commands.
- `harness/checklists/contract-verification.md` is the detailed checklist for
  requirement-led smoke tests and final review.

Typical use:

```bash
../harness/bin/repair-verify repo
../harness/bin/repair-verify repo -- go test ./lib/events ./lib/service
../harness/bin/repair-verify repo --clean-pycache -- python -m compileall -q lib
```

If project dependencies are unavailable, still write a tiny direct smoke with
local stubs for the changed layer. The smoke should execute the contract shape
that hidden tests are likely to assert.
\end{promptbox}

\begin{promptbox}[lst:art-swe-checklist]{SWE-Bench Pro harness: \texttt{checklists/contract-verification.md} (skill)}
# Contract Verification Checklist

Use this as a final gate for each repair. High-severity misses in prior runs
came from not executing the exact prompt contract.

## Requirement Ledger

- Copy exact interface names from the prompt: type/function/method names,
  struct fields, route paths, config keys, defaults, error strings, and payload
  shapes.
- Mark each requirement with one of: implemented, intentionally unchanged with
  reason, or blocked with observed evidence.
- Re-check literal wording before finalizing. Hidden tests often compile
  against exact names such as `StreamEmitter`, `getRaw`, `BackoffDuration`, or
  `durationSeconds`.

## Behavioral Smoke Tests

- Exercise every new public entry point directly, not only syntax or import
  checks. For web/API work, call the controller or route layer and assert the
  response body shape and error translation.
- Cover negative paths from the prompt: missing records, denied privileges,
  full queues, closed emitters, deleted resources, unsupported platforms,
  expired deadlines, failed backends, and boundary positions.
- For state machines, assert transitions at start, middle, end, stopped,
  paused/playing, and invalid input boundaries.
- For data import/parsing tasks, compare complete real fixture outputs against
  expected outputs. Synthetic examples are useful only after at least one real
  regression fixture passes exactly.
- For filesystem/data-cleanup tasks, trace generated or transformed artifacts.
  If the stored record is normalized, test the visible/generated filename too.
- For configurable behavior, test the runtime config value shape used by the
  project, including string values like `"0"` and `"1"` when applicable.
- For multi-backend features, smoke every named backend/path and every explicit
  selector value. Do not validate only the backend that is easiest to run.

## Language/Repo Recipes

- Go: search for `/tmp/go/bin/go` and `/tmp/go/bin/gofmt` before declaring Go
  unavailable. Run gofmt on touched files, focused package tests, and compile
  cross-platform build-tag paths when the prompt mentions a platform.
- Python: run `python -m py_compile` or `compileall` on touched files, then a
  direct smoke for changed behavior. Remove `__pycache__`, `.pyc`, `.pytest_cache`,
  and test environment artifacts before finalizing.
- JavaScript/TypeScript: syntax/type checks do not catch missing runtime imports
  or wrong response shapes. Add a small Node smoke with stubs for changed app,
  controller, or model methods when full dependencies are missing.
- Missing-file tasks: search local/upstream history for the target file and its
  tests. Run the narrow package or upstream tests after adapting the file.

## Final Diff Hygiene

- Run `git diff --check`.
- Run `git status --short --untracked-files=all` and inspect untracked files.
- Remove generated caches, local smoke files, downloaded fixtures, virtualenvs,
  node caches, coverage output, and temporary logs unless the prompt explicitly
  asks to add them.
- If any normal test cannot run, state the exact command and the observed reason,
  then compensate with the closest direct smoke you can execute.
\end{promptbox}

\begin{promptbox}[lst:art-swe-verify]{SWE-Bench Pro harness: \texttt{bin/repair-verify} (tool)}
#!/usr/bin/env bash
set -euo pipefail

usage() {
  cat <<'EOF'
Usage: repair-verify [repo] [--clean-pycache] [-- extra command...]

Runs cheap, task-agnostic verification helpers from inside a SWE-bench repo:
  - discovers common non-PATH Go/Python/Node tools
  - runs git diff --check when the repo is a git worktree
  - runs cheap syntax/package checks when obvious source files exist
  - reports generated artifacts that should usually be removed
  - optionally runs one extra task-specific command

Examples:
  harness/bin/repair-verify repo
  harness/bin/repair-verify repo -- go test ./lib/events ./lib/service
  harness/bin/repair-verify repo --clean-pycache -- python -m compileall -q lib
EOF
}

repo="repo"
clean_pycache=0

while [[ $# -gt 0 ]]; do
  case "$1" in
    --help|-h)
      usage
      exit 0
      ;;
    --clean-pycache)
      clean_pycache=1
      shift
      ;;
    --)
      shift
      break
      ;;
    -*)
      echo "unknown option: $1" >&2
      usage >&2
      exit 2
      ;;
    *)
      repo="$1"
      shift
      ;;
  esac
done

if [[ ! -d "$repo" ]]; then
  echo "repo directory not found: $repo" >&2
  exit 2
fi

cd "$repo"

find_tool() {
  local name="$1"
  shift
  if command -v "$name" >/dev/null 2>&1; then
    command -v "$name"
    return 0
  fi
  for candidate in "$@"; do
    if [[ -x "$candidate" ]]; then
      printf '%s\n' "$candidate"
      return 0
    fi
  done
  return 1
}

go_bin="$(find_tool go /tmp/go/bin/go || true)"
gofmt_bin="$(find_tool gofmt /tmp/go/bin/gofmt || true)"
python_bin="$(find_tool python3 /usr/bin/python3 /usr/local/bin/python3 || find_tool python /usr/bin/python || true)"
node_bin="$(find_tool node /usr/bin/node /usr/local/bin/node || true)"

echo "== discovered tools =="
echo "go: ${go_bin:-not found}"
echo "gofmt: ${gofmt_bin:-not found}"
echo "python: ${python_bin:-not found}"
echo "node: ${node_bin:-not found}"

if git rev-parse --is-inside-work-tree >/dev/null 2>&1; then
  echo
  echo "== git diff check =="
  git diff --check
fi

if [[ -n "$gofmt_bin" ]] && find . -path ./.git -prune -o -name '*.go' -print -quit | grep -q .; then
  echo
  echo "== gofmt check =="
  unformatted="$("$gofmt_bin" -l $(find . -path ./.git -prune -o -name '*.go' -print))"
  if [[ -n "$unformatted" ]]; then
    echo "$unformatted"
    echo "gofmt check failed: run gofmt on the files above" >&2
    exit 1
  fi
fi

if [[ -n "$go_bin" && -f go.mod ]]; then
  echo
  echo "== go package list =="
  "$go_bin" list ./... >/dev/null
fi

if [[ -n "$python_bin" ]] && find . -path ./.git -prune -o -name '*.py' -print -quit | grep -q .; then
  echo
  echo "== python compile touched files =="
  if git rev-parse --is-inside-work-tree >/dev/null 2>&1; then
    mapfile -t py_files < <(git diff --name-only --diff-filter=ACMR -- '*.py')
    if [[ ${#py_files[@]} -gt 0 ]]; then
      "$python_bin" -m py_compile "${py_files[@]}"
    else
      echo "no changed Python files"
    fi
  else
    echo "not a git worktree; skipping touched-file compile"
  fi
fi

if [[ -n "$node_bin" ]] && git rev-parse --is-inside-work-tree >/dev/null 2>&1; then
  mapfile -t js_files < <(git diff --name-only --diff-filter=ACMR -- '*.js' '*.cjs' '*.mjs')
  if [[ ${#js_files[@]} -gt 0 ]]; then
    echo
    echo "== node syntax touched js files =="
    for file in "${js_files[@]}"; do
      "$node_bin" --check "$file"
    done
  fi
fi

if [[ $# -gt 0 ]]; then
  echo
  echo "== extra command =="
  "$@"
fi

if [[ "$clean_pycache" -eq 1 ]]; then
  echo
  echo "== cleaning python caches =="
  find . \( -type d -name __pycache__ -o -type d -name .pytest_cache \) -prune -exec rm -rf {} +
  find . -type f \( -name '*.pyc' -o -name '*.pyo' \) -delete
fi

echo
echo "== generated artifact scan =="
find . \
  -path ./.git -prune -o \
  \( -type d \( -name __pycache__ -o -name .pytest_cache -o -name .mypy_cache -o -name .tox -o -name node_modules -o -name .nyc_output -o -name coverage \) \
  -o -type f \( -name '*.pyc' -o -name '*.pyo' -o -name '*.profraw' -o -name '*.test' \) \) \
  -print | sed -n 's#^\./##;1,200p'

if git rev-parse --is-inside-work-tree >/dev/null 2>&1; then
  echo
  echo "== worktree status =="
  git status --short --untracked-files=all
fi

echo
echo "Reminder: run a direct smoke for the prompt contract: public API shape, negative paths, named backends/platforms, and boundary cases."
\end{promptbox}

\subsection{Terminal-Bench~2}

\begin{promptbox}[lst:art-tb2-readme]{Terminal-Bench~2 harness: \texttt{README.md} (instructions)}
# Terminal-Bench Harness Playbook

Use this harness before solving a task and again before the final answer. The goal is to solve the task on its stated merits, leave the requested artifacts in the container, and avoid wasting time on paths that only satisfy a visible symptom.

## First Pass

1. Extract the contract from the prompt:
   - final artifacts and exact paths, usually under `/app`
   - exact command(s) or invocation style the grader may use
   - allowed and forbidden files; treat `test/` and `tests/` as different names
   - whether the task is black-box/API-only, reverse engineering, package repair, system setup, or stochastic implementation
2. Use the host-to-container script pattern from the prompt for nontrivial commands. It avoids quoting failures and leaves a useful command trail in `/host-ws`.
3. Inspect only sources that are consistent with the task. "Readable" is not the same as "methodologically allowed."
4. Start with a tiny reproducible probe, then scale to the full input. Save intermediate scripts, logs, and validation commands in the host workspace when useful.

## Integrity Boundaries

For tasks framed as black-box recovery, model extraction, API querying, or "by making queries":

- Do not read implementation internals to recover hidden parameters, constants, labels, or answers.
- Import or execute only the public API needed for queries unless the prompt explicitly asks for source inspection.
- Validate recovery with query-derived evidence: held-out probes, recurrence/support diagnostics, round-trip checks, or output agreement. Do not compare against hidden ground-truth variables even if they are exposed.

For tasks asking for an emulator, interpreter, compiler, sampler, server, or independent reimplementation:

- Prefer a general implementation path over task-specific state injection or host-side artifact synthesis.
- If host hooks are necessary, keep them minimal and document why they preserve the requested abstraction.
- Verify the final artifact is produced through the intended path. Example: an emulated program should create its own output through emulated syscalls, not through a host reconstruction of internal memory.

## Final Validation Gate

Before finishing, run a grader-shaped validation:

- Run the exact command from the prompt, with the same flags and paths.
- Compare all observable outputs that matter: files, stdout, stderr, exit status, sizes, and checksums when applicable.
- Verify artifacts are in the container filesystem path that will be graded, not only in `/host-ws`.
- Confirm forbidden paths and grading assets were not used.
- Add at least one hidden-surface smoke check when the visible example is narrow: import more modules, exercise a second API form, test join/post/leave instead of only setup, or check geometry constraints after serialization.

## Package And Container Repair

Use this for Python/R/Cython/library compatibility tasks.

- Keep dependency versions required by the prompt stable. Watch for `pip install` or build tools silently upgrading NumPy or other pinned packages.
- Run the requested snippet from outside the source tree so imports come from the installed package.
- Assert compiled extensions load from compiled artifacts, not fallback `.py` files:

```sh
python - <<'PY'
import inspect
from package import module
print(inspect.getfile(module))
PY
```
- Add broad, non-grader checks after the exact snippet:
  - import representative public modules excluding forbidden test/grader paths
  - `compileall` package code excluding test directories when syntax compatibility is a risk
  - direct smoke calls for compiled helpers
- Reuse `tools/python_package_smoke.py` for broad import, compiled-extension, and compile checks when it fits the package shape.
- Prefer explicit, file-targeted compatibility patches over broad regex rewrites unless the rewrite is followed by a scan for unintended changes.

## Reverse Engineering And Binary Matching

- Collect behavior first: exit status, stdout, stderr, generated files, hashes, file sizes, and compression limits.
- Rebuild with the exact prompt command. Optimized flags can change floating-point or layout-sensitive output.
- Byte-compare every observable artifact. For generated images or binary files, compare both hashes and targeted structure fields.
- If a source size limit exists, check the exact measurement command from the prompt.

## Large File Editing And Vim Macros

- Infer the transformation from head/tail/random samples and row counts before editing the real file.
- Validate the script shape against the allowed command list and count macro keystrokes.
- Prefer file-wide substitutions for simple regular transformations; per-line macros can be much slower at million-row scale.
- When using Vim replacement strings, test escaping and backreferences on a tiny copy first.
- Use portable timing fallbacks such as `date +%s` if `/usr/bin/time` is unavailable.
- Final check: run headless Vim on a saved copy or the real input, then `cmp` against the expected file.

## Geometry And Segmentation Outputs

For mask conversion or image-geometry tasks:

- Use the required model family and real weights when the prompt requires real segmentation. A randomly initialized checkpoint only validates plumbing.
- Keep the CLI general: no hardcoded demo paths; support the prompt's positional/flag style when ambiguous.
- After writing CSV/JSON outputs, validate the serialized artifact, not just the in-memory masks:
  - row count and non-geometry columns preserved
  - all masks are polylines, closed when the format expects closure
  - no rectangle-like fallback polygons
  - exactly one connected component per object
  - no rasterized overlaps after contour export
- Reuse `tools/validate_mask_csv.py` for CSV polygon masks when the column names match the prompt.
- If CPU inference is slow, batch independent model calls where the library supports it.

## Metacircular Or Language Tasks

- Treat visible `test/` programs as sanctioned examples unless the prompt forbids that exact path. Do not touch `tests/`, `test_outputs`, grader logs, or solution files.
- Compare native interpreter output against the new evaluator/interpreter on every visible program.
- Include the prompt's nested/self-hosting example as a final smoke check.
- Avoid special-casing the evaluator's own filename unless the prompt explicitly allows it; it weakens evidence that self-interpretation really works.

## System Service Tasks

- Prefer normal service CLIs over low-level database/API object creation, because defaults often populate domains, styles, policies, queues, and pipelines.
- Verify ownership and runtime directories before starting daemons.
- Do not trust a single status command; also check processes, queues, logs, and direct behavior.
- Build a full behavioral smoke test from the prompt. For a mailing list, that means join confirmation, posting delivery to subscribers, leave confirmation, queue drain, and final policy values.

## Scientific Or Stochastic Implementations

- Confirm the runtime exists and install the minimal required runtime if it does not.
- The public `test()` function should print clear `NAME: PASS` or `NAME: FAIL` lines plus relevant statistics.
- Test both the public sampler and any complex helper modules.
- Include deterministic invalid-input checks and at least one rejection case for violated assumptions, such as non-log-concavity.
- For stochastic outputs, compare distribution statistics with tolerances and persist the required sample file.

## Constraint/Solver Recovery Tasks

- For small ciphers or exact finite-state systems, encode the exact operations as constraints when that is simpler than deriving a full attack by hand.
- Constrain all available known examples when feasible, not just the minimum needed to get one candidate.
- Validate recovered parameters against every known pair, then validate the final deliverable with the provided encrypt/decrypt or round-trip path.
\end{promptbox}

\begin{promptbox}[lst:art-tb2-smoke]{Terminal-Bench~2 harness: \texttt{tools/python\_package\_smoke.py} (tool)}
#!/usr/bin/env python3
"""Run installed-package smoke checks without touching test/grader paths."""

from __future__ import annotations

import argparse
import compileall
import importlib
import inspect
import pkgutil
from pathlib import Path


FORBIDDEN_PARTS = {
    "test",
    "tests",
    "test_outputs",
    "solution",
    "logs",
    "verifier",
    "__pycache__",
}


def forbidden(path: Path) -> bool:
    return any(part in FORBIDDEN_PARTS for part in path.parts)


def iter_modules(package_name: str, limit: int | None) -> list[str]:
    package = importlib.import_module(package_name)
    names = [package_name]
    package_path = getattr(package, "__path__", None)
    if package_path is not None:
        for module in pkgutil.walk_packages(package_path, prefix=f"{package_name}."):
            lower_parts = set(module.name.lower().split("."))
            if lower_parts & FORBIDDEN_PARTS:
                continue
            names.append(module.name)
            if limit is not None and len(names) >= limit:
                break
    return names


def main() -> int:
    parser = argparse.ArgumentParser()
    parser.add_argument("package")
    parser.add_argument("--compiled-module", action="append", default=[])
    parser.add_argument("--import-limit", type=int, default=200)
    parser.add_argument("--compileall", action="store_true")
    args = parser.parse_args()

    failures: list[str] = []
    imported = 0
    for name in iter_modules(args.package, args.import_limit):
        try:
            importlib.import_module(name)
            imported += 1
        except Exception as exc:  # noqa: BLE001 - smoke tool should report all failures
            failures.append(f"import failed {name}: {type(exc).__name__}: {exc}")

    for name in args.compiled_module:
        try:
            module = importlib.import_module(name)
            path = inspect.getfile(module)
        except Exception as exc:  # noqa: BLE001
            failures.append(f"compiled module import failed {name}: {type(exc).__name__}: {exc}")
            continue
        if not any(path.endswith(suffix) for suffix in (".so", ".pyd", ".dll", ".dylib")):
            failures.append(f"{name} did not load from a compiled extension: {path}")

    if args.compileall:
        package = importlib.import_module(args.package)
        roots = [Path(path) for path in getattr(package, "__path__", [])]
        for root in roots:
            for py_file in root.rglob("*.py"):
                if forbidden(py_file):
                    continue
                ok = compileall.compile_file(str(py_file), quiet=1)
                if not ok:
                    failures.append(f"compile failed {py_file}")

    if failures:
        print(f"PYTHON_PACKAGE_SMOKE: FAIL imported={imported}")
        for failure in failures[:80]:
            print(failure)
        if len(failures) > 80:
            print(f"... {len(failures) - 80} more failures")
        return 1

    print(f"PYTHON_PACKAGE_SMOKE: PASS imported={imported}")
    return 0


if __name__ == "__main__":
    raise SystemExit(main())
\end{promptbox}

\begin{promptbox}[lst:art-tb2-mask]{Terminal-Bench~2 harness: \texttt{tools/validate\_mask\_csv.py} (tool)}
#!/usr/bin/env python3
"""Validate polygon mask CSV outputs for segmentation/refinement tasks.

The checker is intentionally structural: it does not judge segmentation quality,
but it catches common grading-surface failures after serialization.
"""

from __future__ import annotations

import argparse
import ast
from pathlib import Path

import numpy as np
import pandas as pd


def parse_coords(value: object) -> list[float]:
    if isinstance(value, (list, tuple, np.ndarray)):
        return [float(x) for x in value]
    text = str(value).strip()
    if not text:
        return []
    try:
        parsed = ast.literal_eval(text)
    except (ValueError, SyntaxError):
        text = text.strip("[]")
        if not text:
            return []
        return [float(part) for part in text.replace(";", ",").split(",") if part.strip()]
    if not isinstance(parsed, (list, tuple)):
        return []
    return [float(x) for x in parsed]


def is_rectangle_like(xs: list[float], ys: list[float]) -> bool:
    points = list(zip(xs, ys))
    if len(points) < 4:
        return False
    unique = list(dict.fromkeys(points))
    if len(unique) == 5 and unique[0] == unique[-1]:
        unique = unique[:-1]
    if len(unique) != 4:
        return False
    return len({x for x, _ in unique}) == 2 and len({y for _, y in unique}) == 2


def rasterize_with_numpy(points: np.ndarray, height: int, width: int) -> np.ndarray:
    mask = np.zeros((height, width), dtype=np.uint8)
    min_x = max(0, int(np.floor(points[:, 0].min())))
    max_x = min(width - 1, int(np.ceil(points[:, 0].max())))
    min_y = max(0, int(np.floor(points[:, 1].min())))
    max_y = min(height - 1, int(np.ceil(points[:, 1].max())))
    if min_x > max_x or min_y > max_y:
        return mask

    x_centers = np.arange(min_x, max_x + 1, dtype=float) + 0.5
    for y in range(min_y, max_y + 1):
        y_center = y + 0.5
        inside = np.zeros_like(x_centers, dtype=bool)
        j = len(points) - 1
        for i in range(len(points)):
            xi, yi = points[i]
            xj, yj = points[j]
            crosses = (yi > y_center) != (yj > y_center)
            if crosses:
                x_intersect = (xj - xi) * (y_center - yi) / (yj - yi) + xi
                inside ^= x_centers < x_intersect
            j = i
        mask[y, min_x : max_x + 1] = inside.astype(np.uint8)
    return mask


def rasterize(xs: list[float], ys: list[float], height: int, width: int) -> np.ndarray:
    pts = np.array([[round(x), round(y)] for x, y in zip(xs, ys)], dtype=np.int32)
    pts[:, 0] = np.clip(pts[:, 0], 0, width - 1)
    pts[:, 1] = np.clip(pts[:, 1], 0, height - 1)
    if len(pts) < 3:
        return np.zeros((height, width), dtype=np.uint8)
    try:
        import cv2

        mask = np.zeros((height, width), dtype=np.uint8)
        cv2.fillPoly(mask, [pts], 1)
        return mask
    except ImportError:
        return rasterize_with_numpy(pts.astype(float), height, width)


def connected_components(mask: np.ndarray) -> int:
    try:
        import cv2

        count, _ = cv2.connectedComponents(mask.astype(np.uint8), connectivity=8)
        return int(count - 1)
    except ImportError:
        seen = np.zeros(mask.shape, dtype=bool)
        active = mask.astype(bool)
        components = 0
        height, width = mask.shape
        for y, x in np.argwhere(active):
            if seen[y, x]:
                continue
            components += 1
            stack = [(int(y), int(x))]
            seen[y, x] = True
            while stack:
                cy, cx = stack.pop()
                for ny in range(max(0, cy - 1), min(height, cy + 2)):
                    for nx in range(max(0, cx - 1), min(width, cx + 2)):
                        if active[ny, nx] and not seen[ny, nx]:
                            seen[ny, nx] = True
                            stack.append((ny, nx))
        return components


def infer_canvas(df: pd.DataFrame, margin: int) -> tuple[int, int]:
    xmax = int(np.ceil(pd.to_numeric(df["xmax"]).max())) + margin
    ymax = int(np.ceil(pd.to_numeric(df["ymax"]).max())) + margin
    return max(1, ymax), max(1, xmax)


def main() -> int:
    parser = argparse.ArgumentParser()
    parser.add_argument("--input-csv", help="Optional original CSV for row/column preservation checks")
    parser.add_argument("--output-csv", required=True)
    parser.add_argument("--width", type=int)
    parser.add_argument("--height", type=int)
    parser.add_argument("--margin", type=int, default=3)
    args = parser.parse_args()

    output_path = Path(args.output_csv)
    df = pd.read_csv(output_path)
    required = {"xmin", "xmax", "ymin", "ymax", "coords_x", "coords_y"}
    missing = required - set(df.columns)
    failures: list[str] = []
    if missing:
        failures.append(f"missing columns: {sorted(missing)}")
        print("\n".join(failures))
        return 1

    if args.input_csv:
        original = pd.read_csv(args.input_csv)
        if len(original) != len(df):
            failures.append(f"row count changed: input={len(original)} output={len(df)}")
        lost = [col for col in original.columns if col not in df.columns]
        if lost:
            failures.append(f"output lost input columns: {lost}")

    height, width = (
        (args.height, args.width)
        if args.height and args.width
        else infer_canvas(df, args.margin)
    )

    occupied = np.zeros((height, width), dtype=np.uint16)
    for idx, row in df.iterrows():
        xs = parse_coords(row["coords_x"])
        ys = parse_coords(row["coords_y"])
        if len(xs) != len(ys):
            failures.append(f"row {idx}: coords_x/coords_y length mismatch")
            continue
        if len(xs) < 4:
            failures.append(f"row {idx}: fewer than 4 polygon points")
            continue
        if (xs[0], ys[0]) != (xs[-1], ys[-1]):
            failures.append(f"row {idx}: polygon is not closed")
        if is_rectangle_like(xs, ys):
            failures.append(f"row {idx}: rectangle-like polygon")
        mask = rasterize(xs, ys, height, width)
        components = connected_components(mask)
        if components != 1:
            failures.append(f"row {idx}: connected components={components}")
        occupied += mask.astype(np.uint16)

    overlap_pixels = int((occupied > 1).sum())
    if overlap_pixels:
        failures.append(f"raster overlap pixels={overlap_pixels}")

    if failures:
        print("MASK_CSV_VALIDATION: FAIL")
        for failure in failures[:50]:
            print(failure)
        if len(failures) > 50:
            print(f"... {len(failures) - 50} more failures")
        return 1

    print(f"MASK_CSV_VALIDATION: PASS rows={len(df)} canvas={width}x{height}")
    return 0


if __name__ == "__main__":
    raise SystemExit(main())
\end{promptbox}

\subsection{GAIA-2}

\begin{promptbox}[lst:art-gaia-readme]{GAIA-2 harness: \texttt{README.md} (instructions)}
# GAIA-2 Harness Notes

Use this as the first checklist for future GAIA-2 ARE tasks. The recurring failures in prior runs were weak evidence before writes, ambiguous references, exact timing/quoted-text drift, missed mandatory side effects, and noisy tool discovery.

## First Minute

1. Read the user request through AgentUserInterface, not the Codex prompt or `poll` alone:

   ```bash
   python task/tools/are.py call AgentUserInterface get_all_messages --json '{}'
   ```

2. Read `task/tools/catalog.json` for app and function names. Avoid dumping the full `are.py list` output unless the catalog is missing or corrupt.
3. Get the simulated current time before resolving relative dates, windows, "today", "this month", or deadline math:

   ```bash
   python task/tools/are.py call SystemApp get_current_time --json '{}'
   ```

4. For each needed function, inspect its exact schema before calling it:

   ```bash
   python task/tools/are.py schema <AppName> <function_name>
   ```

5. Do not read `task/.gaia2/scenario.json`. Avoid broad `.gaia2_state` or sidecar-log inspection; use app APIs unless normal APIs cannot expose required evidence.

## Request Decomposition

Before writes, make a short private checklist:

- Mandatory writes: every send, create, delete, order, move, booking, or final user reply that must happen.
- Conditional writes: the exact condition, evidence needed, and action if the condition is not met.
- Ambiguous references: any phrase that could match multiple emails, files, people, events, products, or conversations.
- Exact details: quoted strings, punctuation, quantities, service types, addresses, dates, times, folders, and recipients.
- Verification: one read-back call after each write, then `python task/tools/are.py poll` after writes that may trigger environment replies.

Do not let a conditional branch swallow a later mandatory instruction. A sentence like "Then email X that the cab was ordered" is still mandatory unless the user clearly made it conditional.

## Evidence Rules Before Acting

- Never send, move, delete, reply to, order, or book based only on filename, loose search hits, or an inferred heuristic.
- For file tasks, confirm the target by content. If an ARE file read returns empty or lazy-loaded content, use the catalog/schema to find another content/search/read operation and keep searching. If content remains unavailable, treat selection as unresolved instead of guessing from a filename.
- For email tasks, search results may match body text as well as subject. If the user asks for a subject/title, read candidate metadata and manually filter by subject.
- For chats/messages, resolve search result IDs back to participant names before acting.
- For product variants, verify availability, exact variant attributes, quantity, and price/order state after checkout.
- For calendar duration/count comparisons, page all events in the relevant range and compute totals once in a small script instead of eyeballing partial views.

## Ambiguity Policy

When a user reference has multiple plausible targets:

1. Enumerate the candidates with stable identifiers and distinguishing metadata.
2. If the user told you to clarify on issues, ask through AgentUserInterface and wait only as long as the task gives you a deadline or the environment convention requires.
3. For irreversible external actions such as sending email, deleting files/events, ordering products, or booking rides, do not invent a fallback after no clarification. Complete unambiguous parts, report the unresolved blocker to the user through AgentUserInterface, and do not perform the ambiguous write.
4. Only use a fallback when the user specified it or the app data gives a single unambiguous, documented rule.

## Timing And Waiting

Treat timing constraints as first-class requirements.

- Compute absolute deadlines from the user message timestamp/current simulated time before doing other work.
- Keep exact target times visible in notes, for example "reply at 2024-10-15 07:04:00; add to cart at 07:04:30".
- Pre-resolve all target IDs, payloads, product variants, and quoted strings before waiting for a deadline. After the wait, execute the deadline writes directly with no extra search or schema discovery in between.
- If several writes share one exact timestamp, use the fastest direct-call path available. Avoid per-item searching between writes; prepare the call arguments first.
- `wait` and notification calls may return early or advance simulated time in chunks. Do not run background or wall-clock loops that keep advancing time while you are acting.
- Prefer bounded waits to the next absolute deadline, then immediately perform the deadline action.
- When monitoring a window, filter/read messages by timestamp and ignore messages outside the requested window, even if they appear during an extra check.
- For post-email replies, verify via timestamp, parent/thread id, sender, and full inbox pagination when needed.

Exact quoted text matters. Preserve punctuation inside quotes, including trailing commas, and verify created titles/bodies by reading them back.

## Pagination And Aggregation

Many list APIs are paginated or have view limits. Page until exhaustion, not until the first useful hit:

- Keep the same filters while advancing offset/page tokens.
- Stop when an empty page is returned or a page has fewer items than the requested limit.
- Record count evidence before the final answer when the task asks for "most", "how many", "latest", or "which app has more".
- Use `harness/are_helper.py page ...` when a function uses simple `limit`/`offset` arguments.

## User-Facing Protocol

- The Codex final answer is only a run summary. It does not count as the GAIA-2 user reply.
- Send progress/completion/blocker messages with AgentUserInterface `send_message_to_user`.
- Before the final AUI message, confirm every mandatory write is either done and verified, intentionally skipped because its condition was false, or blocked by explicit unresolved ambiguity.

## Helper Script

`harness/are_helper.py` wraps common ARE calls and avoids shell quoting/JQ mistakes.

Examples:

```bash
python harness/are_helper.py catalog --app Emails
python harness/are_helper.py schema Emails search_emails
python harness/are_helper.py call AgentUserInterface send_message_to_user --json '{"content":"Done"}'
python harness/are_helper.py page Contacts list_contacts --limit 100 --json '{}'
```

Use the helper for convenience, but still inspect the schema for exact argument names.
\end{promptbox}

\begin{promptbox}[lst:art-gaia-helper]{GAIA-2 harness: \texttt{are\_helper.py} (tool)}
#!/usr/bin/env python3
"""Small ARE command helper for GAIA-2 tasks.

The script intentionally stays generic: it shells out to task/tools/are.py,
keeps JSON quoting inside Python, and provides a simple offset/limit paginator
for APIs with standard pagination arguments.
"""

from __future__ import annotations

import argparse
import json
import subprocess
import sys
from pathlib import Path
from typing import Any


COMMON_ITEM_KEYS = (
    "items",
    "results",
    "data",
    "records",
    "messages",
    "emails",
    "events",
    "contacts",
    "conversations",
    "products",
    "orders",
    "rides",
    "files",
    "entries",
)


def parse_json_arg(raw: str) -> dict[str, Any]:
    try:
        value = json.loads(raw)
    except json.JSONDecodeError as exc:
        raise SystemExit(f"--json is not valid JSON: {exc}") from exc
    if not isinstance(value, dict):
        raise SystemExit("--json must decode to an object")
    return value


def are_py(task_root: str) -> Path:
    path = Path(task_root) / "tools" / "are.py"
    if not path.exists():
        raise SystemExit(f"ARE tool not found: {path}")
    return path


def run_are(task_root: str, args: list[str]) -> Any:
    cmd = [sys.executable, str(are_py(task_root)), *args]
    proc = subprocess.run(cmd, text=True, capture_output=True)
    if proc.returncode != 0:
        sys.stderr.write(proc.stderr)
        sys.stderr.write(proc.stdout)
        raise SystemExit(proc.returncode)

    out = proc.stdout.strip()
    if not out:
        return None
    try:
        return json.loads(out)
    except json.JSONDecodeError:
        return out


def print_json(value: Any) -> None:
    print(json.dumps(value, indent=2, ensure_ascii=False, sort_keys=True))


def command_catalog(args: argparse.Namespace) -> None:
    path = Path(args.task_root) / "tools" / "catalog.json"
    if not path.exists():
        raise SystemExit(f"Catalog not found: {path}")
    catalog = json.loads(path.read_text())
    if args.app is None:
        print_json(catalog)
        return

    needle = args.app.lower()

    def keep(value: Any) -> Any:
        if isinstance(value, dict):
            kept: dict[str, Any] = {}
            for key, child in value.items():
                child_kept = keep(child)
                if needle in str(key).lower() or child_kept not in (None, {}, []):
                    kept[key] = child if needle in str(key).lower() else child_kept
            return kept
        if isinstance(value, list):
            kept_list = [item for item in (keep(item) for item in value) if item not in (None, {}, [])]
            return kept_list
        if needle in str(value).lower():
            return value
        return None

    print_json(keep(catalog))


def command_schema(args: argparse.Namespace) -> None:
    print_json(run_are(args.task_root, ["schema", args.app, args.function]))


def command_call(args: argparse.Namespace) -> None:
    payload = parse_json_arg(args.json)
    print_json(
        run_are(
            args.task_root,
            ["call", args.app, args.function, "--json", json.dumps(payload, ensure_ascii=False)],
        )
    )


def extract_items(result: Any, explicit_key: str | None) -> list[Any]:
    if explicit_key:
        if not isinstance(result, dict) or explicit_key not in result:
            raise SystemExit(f"Result does not contain items key {explicit_key!r}")
        items = result[explicit_key]
    elif isinstance(result, list):
        items = result
    elif isinstance(result, dict):
        items = None
        for key in COMMON_ITEM_KEYS:
            if isinstance(result.get(key), list):
                items = result[key]
                break
        if items is None:
            list_values = [value for value in result.values() if isinstance(value, list)]
            if len(list_values) == 1:
                items = list_values[0]
            else:
                raise SystemExit(
                    "Could not infer list field. Re-run with --items-key. "
                    f"Top-level keys: {', '.join(result.keys())}"
                )
    else:
        raise SystemExit(f"Cannot paginate non-list result of type {type(result).__name__}")

    if not isinstance(items, list):
        raise SystemExit("Inferred items value is not a list")
    return items


def command_page(args: argparse.Namespace) -> None:
    base = parse_json_arg(args.json)
    offset = args.start
    all_items: list[Any] = []
    pages = 0

    while True:
        payload = dict(base)
        payload[args.limit_key] = args.limit
        payload[args.offset_key] = offset
        result = run_are(
            args.task_root,
            ["call", args.app, args.function, "--json", json.dumps(payload, ensure_ascii=False)],
        )
        items = extract_items(result, args.items_key)
        all_items.extend(items)
        pages += 1
        if not items or len(items) < args.limit:
            break
        if args.max_pages is not None and pages >= args.max_pages:
            break
        offset += args.limit

    print_json({"count": len(all_items), "pages": pages, "items": all_items})


def build_parser() -> argparse.ArgumentParser:
    parser = argparse.ArgumentParser(description=__doc__)
    parser.add_argument("--task-root", default="task", help="Task root containing tools/are.py")
    sub = parser.add_subparsers(dest="command", required=True)

    catalog = sub.add_parser("catalog", help="Print catalog.json, optionally filtered by app text")
    catalog.add_argument("--app")
    catalog.set_defaults(func=command_catalog)

    schema = sub.add_parser("schema", help="Print one function schema")
    schema.add_argument("app")
    schema.add_argument("function")
    schema.set_defaults(func=command_schema)

    call = sub.add_parser("call", help="Call one ARE function with JSON args")
    call.add_argument("app")
    call.add_argument("function")
    call.add_argument("--json", default="{}")
    call.set_defaults(func=command_call)

    page = sub.add_parser("page", help="Collect all pages for offset/limit list APIs")
    page.add_argument("app")
    page.add_argument("function")
    page.add_argument("--json", default="{}", help="Base JSON args excluding limit/offset")
    page.add_argument("--limit", type=int, default=100)
    page.add_argument("--start", type=int, default=0)
    page.add_argument("--limit-key", default="limit")
    page.add_argument("--offset-key", default="offset")
    page.add_argument("--items-key")
    page.add_argument("--max-pages", type=int)
    page.set_defaults(func=command_page)
    return parser


def main() -> None:
    parser = build_parser()
    args = parser.parse_args()
    args.func(args)


if __name__ == "__main__":
    main()
\end{promptbox}

\end{document}